\documentclass{article}
\usepackage{jfrExamplee}
\usepackage{graphicx}
\usepackage{apalike}
\usepackage{setspace}
\usepackage[cmex10]{amsmath}
\usepackage{amssymb}
\usepackage{algorithm}
\usepackage{gensymb}
\usepackage{algpseudocode}
\usepackage{graphicx}
\usepackage{array}
\usepackage{mdwmath}
\usepackage{mdwtab}
\usepackage{amsmath}
\usepackage{amssymb,amsmath}
\usepackage{multirow}
\usepackage{xcolor}
\usepackage{caption}
\usepackage{booktabs}
\usepackage[framemethod=tikz]{mdframed}
\usetikzlibrary{calc}
\usepackage{lipsum}
\usepackage{tabularx}
\usepackage{booktabs}
\usepackage{float}
\usepackage[caption=false,font=footnotesize]{subfig}
\usepackage{url}
\usepackage{placeins}
\title{3D Scan Registration using Curvelet Features in Planetary Environments}

\author{
Siddhant Ahuja \\
Department of Mechanical and Mechatronics Engineering\\
University of Waterloo\\
Waterloo, ON, Canada N2L3G1 \\
\texttt{s3ahuja@uwaterloo.ca} \\
\And
Peter Iles \\
Neptec Design Group\\
Kanata, ON, Canada K2K 1Y5 \\
\texttt{pjwiles at neptec.com} \\
\AND
Steven L. Waslander \\
Department of Mechanical and Mechatronics Engineering\\
University of Waterloo\\
Waterloo, ON, Canada N2L3G1 \\
\texttt{stevenw@uwaterloo.ca} \\
}

\begin{document}
\doublespacing
\maketitle

\begin{abstract}
Topographic mapping in planetary environments relies on accurate 3D scan registration methods. However, most global registration algorithms relying on features such as FPFH and Harris-3D show poor alignment accuracy in these settings due to the poor structure of the Mars-like terrain and variable resolution, occluded, sparse range data that is hard to register without some a-priori knowledge of the environment. In this paper, we propose an alternative approach to 3D scan registration using the curvelet transform that performs multi-resolution geometric analysis to obtain a set of coefficients indexed by scale (coarsest to finest), angle and spatial position. Features are detected in the curvelet domain to take advantage of the directional selectivity of the transform. A descriptor is computed for each feature by calculating the 3D spatial histogram of the image gradients, and nearest neighbor based matching is used to calculate the feature correspondences. Correspondence rejection using Random Sample Consensus identifies inliers, and a locally optimal Singular Value Decomposition-based estimation of the rigid-body transformation aligns the laser scans given the re-projected correspondences in the metric space. Experimental results on a publicly available data-set of planetary analogue indoor facility, as well as simulated and real-world scans from Neptec Design Group's IVIGMS 3D laser rangefinder at the outdoor CSA Mars yard demonstrates improved performance over existing methods in the challenging sparse Mars-like terrain. 
\end{abstract}

\section{Introduction}

\begin{figure}[htb!]
      \centering
      \includegraphics[width=0.9\textwidth]{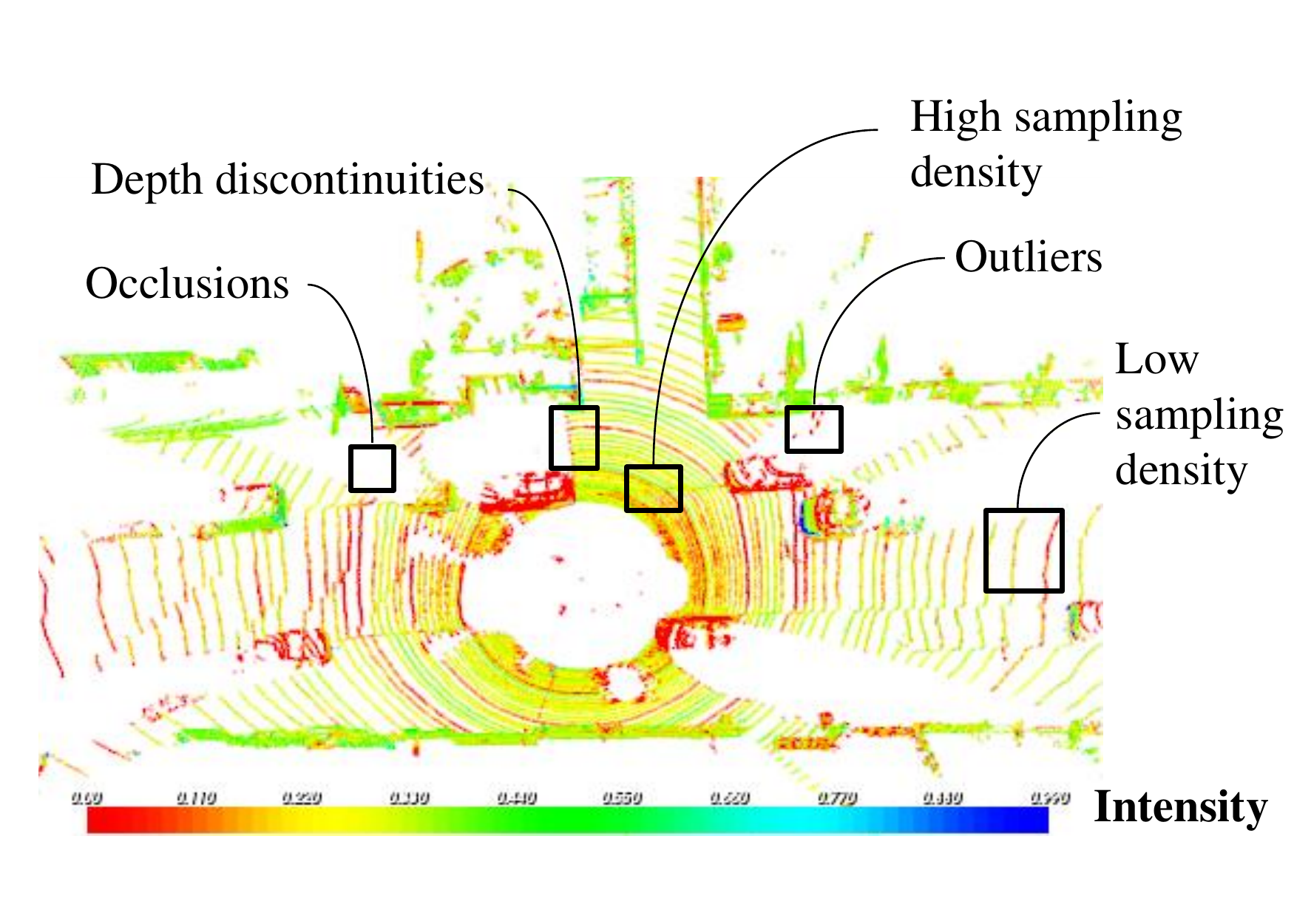}

      \caption{3D-LIDAR intensity scan captured by Velodyne HDL-64E sensor from KITTI data-set \cite{2012-geiger-kitti} highlighting regions with varying sampling density, occlusions, depth-discontinuities and outliers (best viewed in colour). }
      \label{problems_laser_point_cloud}
   \end{figure}

3D mapping of unstructured environments relies on accurate alignment of partially overlapping scans into a globally consistent model, called scan registration. Sensors such as RGB-D cameras, LIDAR, Time of Flight (ToF), and stereo cameras provide information as point-sampled 3D surfaces, termed point-clouds. Overlapping scans share a common set of points that can be used for matching in order to estimate the relative rigid body transformation between scans (6-DOF rotation and translation). Separate views of the same environment can be accumulated into a global coordinate system which helps an intelligent mobile robot perform tasks in an unstructured environment. However, points within each scan represent samples of different surfaces within the environment, subject to the type of sensor used for capturing the scene, sampling density (number of points per volumetric unit), sensor viewpoint (relative geometric position), sensitivity to measurement noise, quantization errors, occlusions, depth-discontinuities due to sharp edges, and the surface characteristics of the objects within the scene such as color, shape, textures, etc. (see Figure \ref{problems_laser_point_cloud}). Finding accurate transformation parameters, given the intra-scan problems and a relatively large initial inter-scan transformation error, makes the registration problem especially hard.

Topographic mapping is one such instance where these issues are prevalent. Desire for the establishment of a permanent presence on extraterrestrial surfaces, detailed and accurate mapping of the terrain and rover localization with respect to the environment is essential to conduct operations such as exploration, site selection, base construction etc. Without an absolute positioning system such as GPS available in space, on-board sensing must be used for localization and mapping. The use of 3D laser scan data for planetary exploration has been proposed by numerous researchers \cite{Tong01042013,Carle2010,hamel2012}, as a means to improve on existing stereo methods in generating detailed 3D map data, localizing over long distances and operating in low ambient light conditions.  These advantages have led agencies to pursue numerous next generation rover technologies with LIDAR sensing at the core of the autonomy packages. 

One of the challenges with most LIDAR sensor configurations is that the needs for rover autonomy are somewhat contradictory: exploration requires rapid update over long ranges, while detailed mapping requires high densities of local points with minimal uncertainty. The poor structure of the Mars like terrain coupled with the shallow grazing angle results in a variable resolution, occluded, sparse range data that is hard to register without some a-priori knowledge of the environment. Due to the lack of natural and man-made features such as trees and buildings, most registration algorithms show poor convergence properties.

The same issues arise in other applications that require dense scan registrations, such as indoor mapping, mining and highway surveys. In order to maintain a safe work environment and increase productivity, automated  LIDAR based surface profiling systems are preferred over traditional acquisition methods. Existing registration methods show poor convergence properties in areas such as indoor spaces, underground and open-pit mines, where the lack or insufficient quality of absolute localization based on GPS, and the drift of relative inertial localization cause a significant problem in mapping. Automated mapping in rural or off-road environments is another example where the sparse scene structure, poor localization and dynamic objects may result in inaccurate laser scan alignments.

In this paper, a novel approach to scan registration using the curvelet transform is presented. Suitable features are extracted from the curvelet domain via difference of curvelets operator at multiple scales followed by extrema detection and filtering. Feature descriptors around the candidate key-points are computed from spatial histograms of image gradients and the correspondences are found using nearest neighbor matching. Feature correspondences are filtered using Random Sample Consensus (RANSAC) to reject outliers and the laser scans are aligned using Singular Value Decomposition (SVD) based estimation of rigid body transformation. This approach is adapted to work with both a vertically scanning SICK LMS291 laser rangefinder and hybrid IVIGMS scanner provided by Neptec Design Group Ltd. Experimental results from three different data-sets representing indoor and outdoor environments, comparing the average root-mean-squared errors in translation and rotation for existing methods as well as proposed approach demonstrate the performance of our algorithm in the challenging sparse Mars-like terrain.

The rest of the paper is organized as follows: Section 2 provides the related work in the area of scan registration and a problem formulation is presented in Section 3. Details of the proposed method are given in Section 4. Quantitative and qualitative results for indoor and outdoor 3D laser scan data-sets are provided in Section 5 with a discussion on suitability of the algorithm for mapping. Section 6 concludes the paper with directions for future work.

\section{Related Work}

One of the most popular scan registration methods, the iterative closest point (ICP) \cite{1992-besl-icp,1991-chen-icp,1994-zhang-icp}, relies on point-to-point correspondences to estimate the relative transformation of scans by minimizing the Euclidean distance error metric. The original ICP algorithm assumes that there exists a correspondence between each point of the source and model data-sets. This assumption is often violated with partially overlapping scans. Some modifications to the ICP algorithm have included the maximum error cutoff metric \cite{1996-masuda-maximum-cutoff-icp} to account for false correspondences, and did not require every point to be matched. One of the key problems with scan registration is that the sparsely sampled corresponding points in two different scans often do not correspond to the same point in the 3D environment, but ICP assumes that they do.  In addition, the quality of the ICP solution depends heavily on the availability of good initial estimates of the transformation \cite{1992-besl-icp}.

Many extensions to the original ICP have been proposed that transform the point clouds from metric space to feature space for fast correspondence based matching. They rely on finding unique features in the two scans, in order to improve the registration accuracy. Features based on color and intensity values \cite{1994-godin-intensity}, normals \cite{2012-ioannou-don,2004-alexa-normals}, curvatures \cite{1996-chua-curvatures}, integral volume descriptors \cite{2005-gelfand-ivd}, moment invariants \cite{1980-sadjadi-moment-invariants}, spherical harmonics \cite{1995-burel-sphericalharmonics}, spin images \cite{1997-johnson-spin-images}, corners, lines and planes \cite{2005-censi-hough}, the scale-invariant feature transform (SIFT) \cite{2012-Henry-RGBD}, and combinations of the above \cite{2002-sharp-invariant-icp} have all been suggested. However, all of these features are prone to measurement noise and cannot deal with varying sampling density within the point cloud. Locally planar surface structure was exploited by Segal et al. \cite{2009-segal-gicp} for plane-to-plane correspondence search in the generalized iterative closest point algorithm (GICP). In order to incorporate local surface characteristics, Bosse et al. modified the correspondence step in ICP to include a nine-dimensional vector consisting of local centroids and two eigen-vectors representing planar and cylindrical regions in a voxelized point cloud  \cite{2009-bosse-continuous-matching,2012-bosse-spring-mounted}. This approach has been shown to work well in scenarios where the vehicle is continuously acquiring 3D scans while in motion. However, due to the vehicle motion and trajectory smoothness constraints, it is unable to handle a large inter-scan transformation error. Point features on sharp edges and planar surface patches have previously been proposed for scan registration \cite{2014-zhang-loam} where the ICP score has been modified to include point-to-line/plane correspondence search with local smoothness constraints. The assumption of a constant velocity motion model limits its applicability to continuous scan matching. In addition, the requirement for planar structures in the scene restricts the applicability of the algorithm to sparse outdoor environments.

Rusu et al. introduced the Sample Consensus-Initial Alignment (SAC-IA) algorithm \cite{2008-rusu-pfh,2009-rusu-fpfh} using 16-dimensional fast point feature histograms (FPFH) that describe the local surface structure. Experimental results showing the robustness of these features to outliers and invariance to pose, sampling density, and measurement noise are lacking in the literature.  Various heuristics based on false correspondence rejection and re-weighting have tried to improve the robustness but the convergence is not guaranteed. In addition, these features require extensive computational steps and the resulting transformation is only an approximation due to the compact representation of a 3D surface (in metric space) as a feature point in feature space.

The Harris corner point detector initially proposed for 2D images \cite{1988-harris3d} has previously been extended for key-point detection on 3D surfaces \cite{2009-steder-narf-robust,2010-sipiran-harris3d} where it makes use of surface normals instead of using image gradients. 3D normals calculated in noisy regions or around points at depth-discontinuities are frequently incorrect.

Normal aligned radial features (NARF) proposed by Steder et al. operates on range images rather than laser scans, similar to the proposed method \cite{2010-Steder-NARF,2011-Steder-NARF}. Object boundaries are computed in depth-discontinuous regions by traversing from foreground to background in the range image, followed by key-point localization in stable surface areas around these edges. This type of feature works well for object recognition where there are large objects away from significant depth-discontinuous regions. Planetary environments often contain smoothly varying surfaces with small rocks and debris and thus, NARF features are typically found near the ground level, making it harder to detect sufficient number of key-points which can better represent the surface variations.

The spin-image algorithm for object recognition and pose estimation first proposed by Johnson et al. for 2D images \cite{1997-johnson-spin-images} was later extended to work with 3D laser scans \cite{1999-johnson-spin-images}. Surface around the key-point is represented as a gray-scale image where darker areas correspond to regions of high point density. Spin-image based descriptors do not explicitly make use of the information outside of the object boundaries, making them less reliable for pose estimation \cite{2009-steder-narf-robust}. In addition, spin image descriptors produce ambiguous correspondence matches in cluttered scenes and noisy environments.

Frequency-domain based approaches decouple the problem of finding rotation and translation transformation parameters and attempt to find a suitable registration in the transformed domain \cite{2013-Heiko-spectral,2002-Lucchese-frequency,2006-Keller-pseudopolar}. Phase correlation is typically employed for matching which is robust to the effects of noise and occlusions, while fast Fourier transforms (FFT) used to compute cross-correlations makes this approach computationally efficient. However, the Fourier transform can only retrieve the global frequency content of the signal and provides a dense representation of the underlying signal.

Another transformation found in the literature relies on finding a translation invariant Fourier transform on two Extended Gaussian Images (EGI) \cite{2006-makaida-egi} of laser scans,  where the surface normals of an object are mapped onto the unit sphere. However, this approach can only be applied to smooth surfaces and fails to match surfaces with constant EGI (such as a sphere). Censi et al. \cite{2009-censi-hsm3d} proposed another approach to scan matching that projects the two scans into the Hough/Radon domain \cite{2009-censi-hsm3d} defined on the unit sphere. Similar to the work of Makaida et al., a translation invariant spectrum is computed to find the rotation and cross-correlations are used to find the translation. Both EGI approaches and the ones based on transformation to Hough/Radon domain are sensitive to the measurement noise and sampling density during the calculation of normals.

Unlike the shape-fixed rectangles in the frequency domain of conventional FFT, multi-scale transforms such as the discrete wavelet transform (DWT) use dilated shape varying rectangles to find directional elements such as edges and ridge features in the laser scans. However, many wavelet coefficients are needed to account for singularities along lines or curves. To overcome this problem, other directional wavelets such as wedgelets \cite{1999-David-wedgelet}, beamlets \cite{2002-barth-beamlet}, contourlets \cite{2005-Do-contourlet}, surfacelets \cite{2007-Lu-surfacelet}, etc. have been proposed, however the detected features are less prominent especially at the edges. In order to account for curve-singularities, the curvelet transform has previously been proposed which generates a sparse representation of the points within the scan and employs angled polar wedges in the frequency domain to find directional features. Previously, Alam et al. applied the curvelet transform for the problem of image fusion \cite{2012-alam-entropy-curvelet-image} where only an approximate sub-band of the coefficients was used for registration. The algorithm relied on the assumption that the curvelet coefficients were normally distributed. 

Depth-discontinuities of the surfaces seen from different perspectives are often distinct, and can be detected using various edge-extraction algorithms. Key-points and descriptors can be calculated around these regions for correspondence estimation between laser scan pairs. However, in planetary environments, typical edge detection methods fail due to the smoothly varying surfaces.  Instead, curvelets can be applied directly to the range image to capture the smoothly varying surface and define reliable features that work well for scan registration. This work has been inspired by the techniques used in the computer vision field for image fusion, and to the best knowledge of the authors, there has not been a detailed study of the application of the curvelet transform to the scan registration problem.

\section{Problem Formulation}

Two 3D point-sets are defined: the model set $M=\{m_1, \cdots, m_{N_{M}\}}$ and the data set $D=\{d_1, \cdots, d_{N_D}\}$ where $m_i, d_j \in \mathbb{R}^3$ for $i\in\{1,\cdots,N_M\}, j\in\{1,\cdots,N_D\}$. A scan-to-scan registration algorithm seeks to identify a 6-DOF transformation of the data scan to match a model scan coordinate frame to form a single, globally consistent model of the environment. This is done by maximizing the similarity between scans after transformation. An estimate, $T$, of the transformation $T^*=\{R,t\} \in \mathbb{SE}(3)$, with rotation $R = \{R_x, R_y, R_z\} \in \mathbb{SO}(3)$ and translation $t = \{t_x, t_y, t_z\} \in \mathbb{R}^3$ can be obtained from:

\begin{equation}
\label{problem_formulation}
T^* = \underset{T\in \mathbb{SE}(3)}{\arg\!\max}\,\, C(M, T(D))
\end{equation}
where $C(M, T(D))$ is the similarity metric between the model set $M$ and the transformed data set $T(D)$.

\section{Proposed Method}

In order to incorporate curvelet transform into the scan registration process, we define a similarity metric that is computed via a five-part algorithm as follows.

\begin{enumerate}
\item Range images $R_M$ and $R_D \in I = \mathbb{R}_+^{X \times Y}$ of dimension $X \times Y$ are constructed from the spherical projections of the 3-D laser scans, $M$ and $D$. Background regions surrounded by a connected border of foreground pixels are assigned an intensity value by employing a hole filling algorithm \cite{2003-soille-book}. A Gaussian filter of size $3 \times 3$ with a standard deviation of 0.5 is used to smooth the range images and is followed by normalization of the range intensity values. Figure~\ref{range_image} shows the range image generated from the first 3D laser scan in the \textit{a100} Mars Dome data-set \cite{2011-chi-dataset}, with an angular resolution of 0.5 degrees in both $x$ and $y$ directions.
\begin{figure}[htb!]
      \centering
      \includegraphics[width=0.6\textwidth]{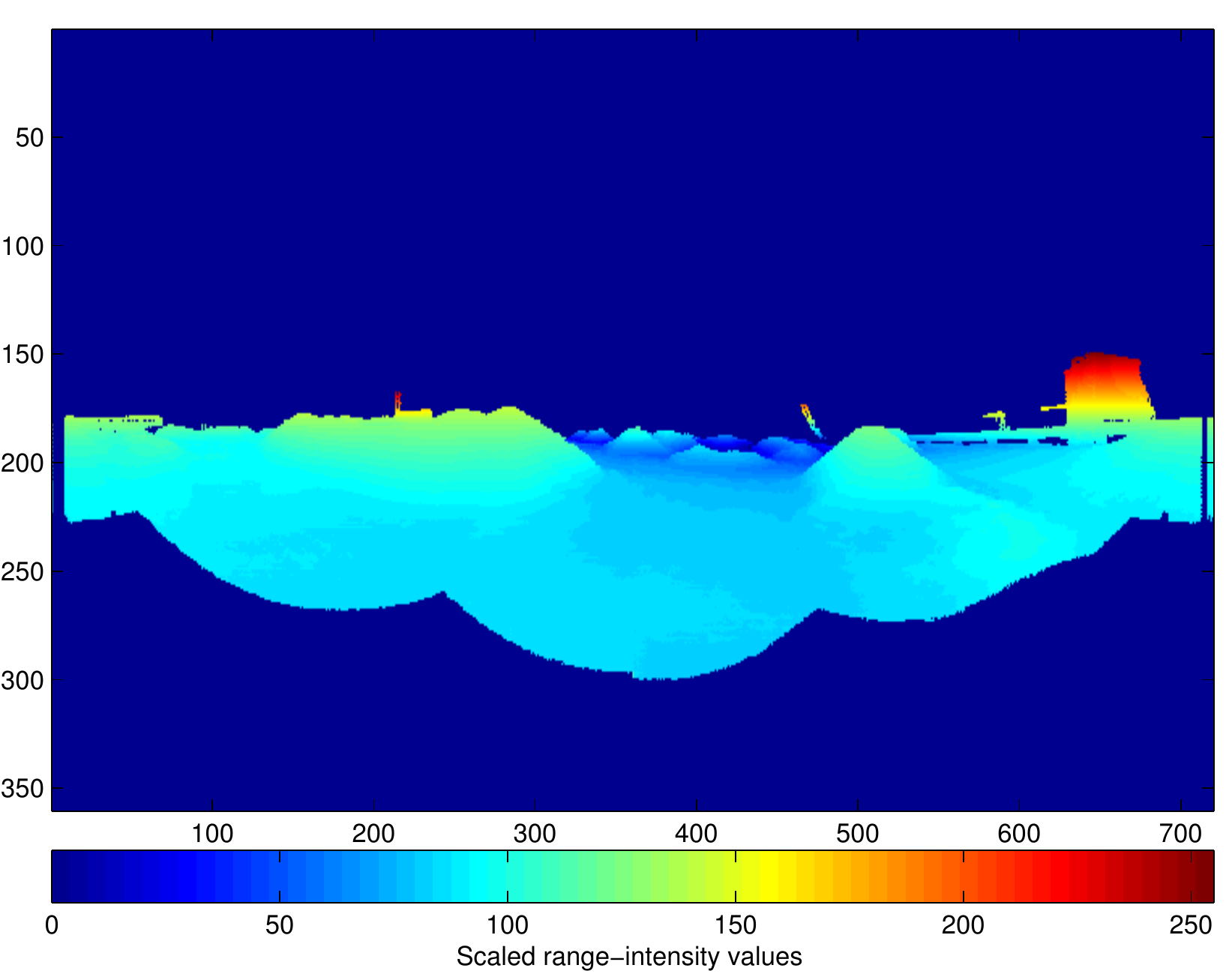}
      \caption{Range image generated from the spherical projection of the first 3D laser scan in the \emph{a100} Mars Dome data-set \cite{2011-chi-dataset}, with an angular resolution of 0.5 degrees in both $x$ and $y$ directions.}
      \label{range_image}
\end{figure}
\item The discrete curvelet transform is then applied to each range image to obtain two sets of curvelet coefficients. The discrete curvelet transform is a linear digital transformation consisting of complex valued basis functions $\Psi_{j, l, k} : \mathbb{R}^2 \rightarrow \mathbb{C}$ parametrized in three spaces: scale $2^{-j} \in \mathbb{R}$, orientation $\theta_l[0,2\pi) = 2\pi\cdot 2^{-\lfloor j/2 \rfloor}\cdot l,$ where  $l=0,1,\ldots 2\lfloor j/2 \rfloor-1 \in \mathbb{Z}$, and scale dependent relative position $x_k^{(j,l)} = R_{\theta_l}^{-1}(k_1 2^{-j},k_2 2^{\lfloor j/2 \rfloor})\in I$, where $k = (k_1,k_2)$ indexes a standard translational grid that is adjusted to each scale value. The notation $\lfloor x \rfloor$ denotes the floor of x, which truncates a positive real number to its integer component. The discrete curvelet transform of an $n \times m$ Cartesian array formed from the range image of size $ X \times Y$ pixels is defined as the inner product between an element of the array $f(t_1,t_2)  0\leq t_1 < n, 0 \leq t_2 < m $, and the curvelet basis function $\Psi_{j, l, k}$, given as \cite{2006-candes-curvelet}:
\begin{eqnarray*}
    c(j,l,k) &=& \langle f, \Psi_{j, l, k} \rangle \\ 
      &=& \sum_{0\leq t_1 < n, 0 \leq t_2< m}f[t_1,t_2] \Psi_{j,l,k}[t_1, t_2]
\end{eqnarray*}
where $\Psi_{j,l,k}$ is the basis function for the discrete version of the forward transform and $c(j,l,k)$ is the indexed curvelet coefficient. The curvelet transform is implemented using second generation fast discrete curvelet transform (FDCT) via wrapping is available at \url{ http://www.curveleab.org}. Figure \ref{fdct_log} presents the log of the curvelet coefficients for the range image in Figure \ref{range_image} for scales from the coarsest to level 4, and for angles from the 2nd coarsest to level 16. The center of the display shows the low frequency coefficients at the coarsest scale, with the Cartesian concentric coronae at the outer edges at various scale levels, showing coefficients at higher frequencies. Each corona contains four strips which are subdivided into angular panels \cite{2006-candes-curvelet}.

\begin{figure}[htb!]
      \centering
      \includegraphics[width=0.6\textwidth]{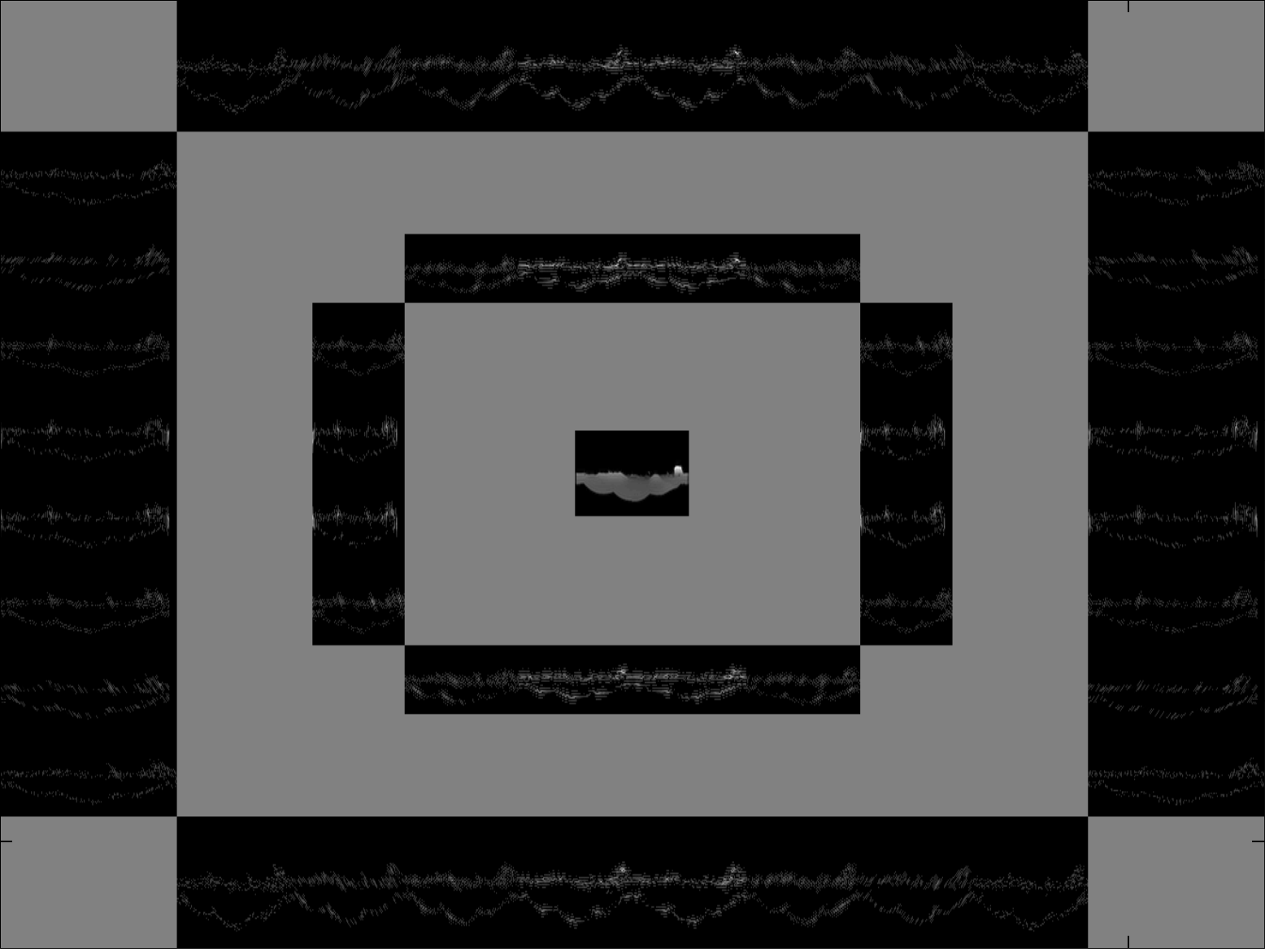}
      \caption{Log of the curvelet coefficients for the range image in Figure \ref{range_image} for $\lambda = 1 \ldots 4$ and $\phi = 2 \ldots 16$.}
      \label{fdct_log}
\end{figure}

\item  The Cartesian array formed from the range image can be reconstructed from the curvelet coefficients $c(j,l,k)$ by taking the inverse curvelet transform as \cite{2006-candes-curvelet}:

\begin{equation}
I_c = \sum_{j,l,k} c(j,l,k) \tilde{\psi}_{j,l,k}
\end{equation}
where $\tilde{\Psi}_{j,l,k}$ is the basis function for inverse transform.  Additionally, it is possible to invert each scale level individually, leading to scale dependent reconstructed images.
\begin{equation}
I_c(j) = \sum_{l,k} c(j,l,k) \tilde{\psi}_{j,l,k}
\end{equation}

A novel differences of curvelets (DoC) image feature is introduced to identify stable locations that are invariant to scale. Contributions from individual sub-bands from two nearby scales are subtracted to produce a set of difference-of-curvelet images as follows:

\begin{equation}
I_{DoC}(j) = I_c(j) - I_c(j-1)
\end{equation}

Similar to the Scale Invariant Feature Transform (SIFT) \cite{2004-lowe-sift}, local maxima and minima over scale and space are used to find potential key-points by comparing the pixel value at the current scale with its 8 connected neighbors, and 9 other pixels in both the previous and next scales. The result is then thresholded to eliminate low-contrast key-points, and key-points lying close to the minimum cut-off range of the sensor, to obtain robust interest points (depicted as red stars in Figure \ref{curvelet_features}). Some of the key-points are found at the interface of data and missing data, whereas others are found at locations where there is a sudden change in range intensity data values within a local neighborhood. The aim of the descriptor is to capture the local surface variations around the key-point such that valid correspondences can be computed at later stages. A 16x16 neighborhood around the key-point is used to obtain a 128 bin feature descriptor from the 3D spatial histogram of image gradients \cite{2004-lowe-sift,2005-Mikolajczyk-comparison-descriptors}.

\begin{figure}[htb!]
      \centering
      \includegraphics[width=0.7\textwidth]{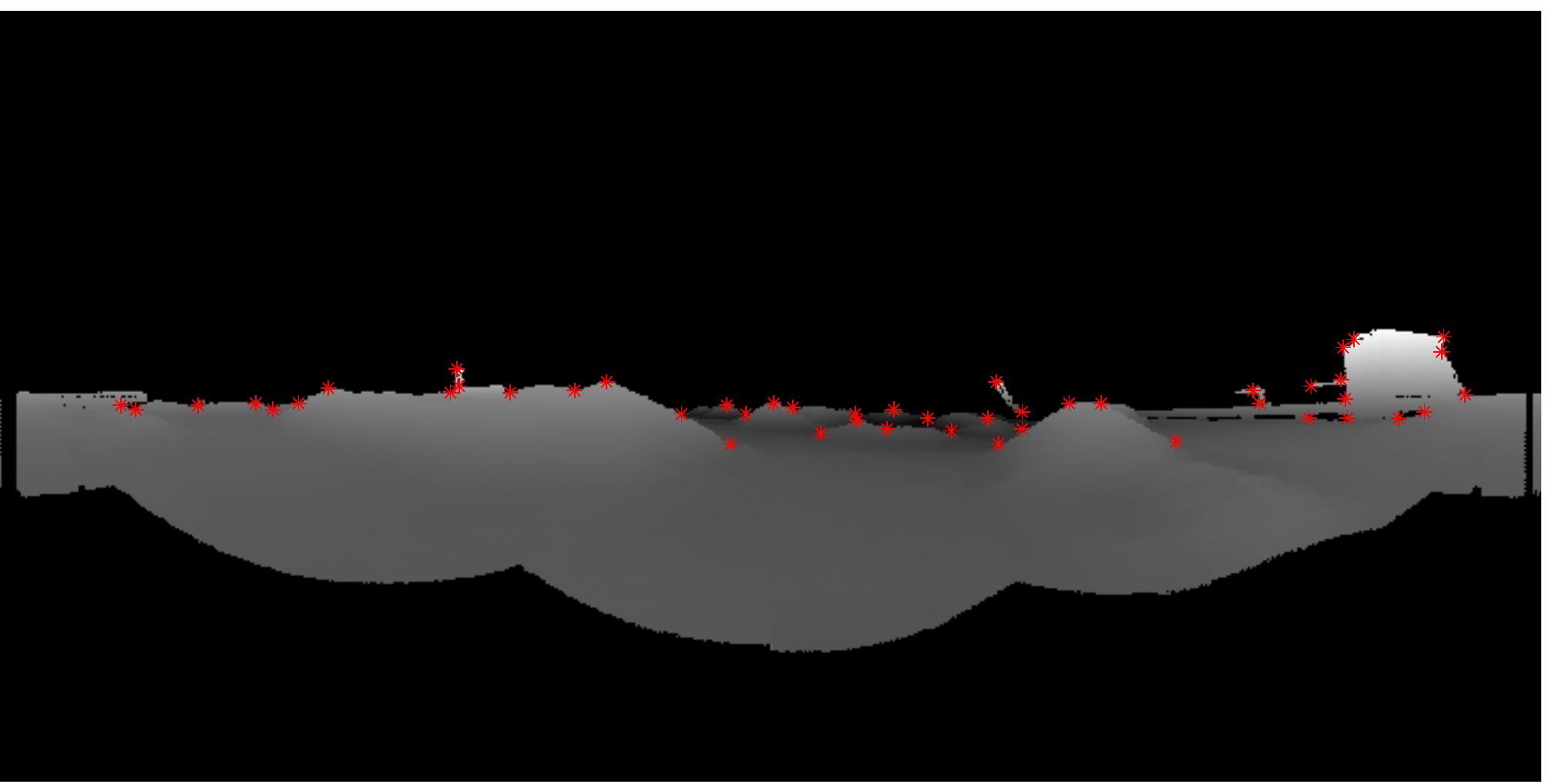}
      \caption{Curvelet features (as red stars ) from the range image in Figure \ref{range_image}.}
      \label{curvelet_features}
\end{figure}

\item A quick and efficient feature matching is performed using approximate nearest neighbor search in the feature space and feature correspondences are established between curvelet feature pairs. The nearest neighbor is defined as the feature with minimum Euclidean distance to another feature in the feature space \cite{muja_flann_2009}. Feature correspondences are filtered using RANSAC to reject outliers \cite{1981-ransac}.

\item Filtered corresponding curvelet feature pairs in the metric space are used to estimate the rigid body transformation by employing SVD \cite{1987-arun-svd}. SVD algorithm provides a closed form least-squares solution for the rotation matrix and translation vector given the correspondence pairs. Other methods such as Levenberg-Marquardt least-squares algorithm \cite{1963-marquardt,1944-Levenberg} provide an iterative solution to the rigid body transformation estimation problem, however the SVD algorithm provides the best possible solution in one step without the need for an initial guess.

\end{enumerate}

\section{Experimental Results}
The proposed approach is evaluated using three planetary analogue 3D LIDAR scan data-sets (The University of Toronto Institute for Aerospace Studies Mars-dome analogue indoor data-set, IVIGMS simulated and real-world CSA Mars-Yard outdoor data-sets) in both indoor and outdoor settings. Registering scans from emulated Mars terrain is quite challenging due to a large inter-scan transformation error and a lack of sufficient features to match the scans with each other, even with full $360^{\circ}$ x $180^{\circ}$ scans of the terrain. The poor structure of the Mars like terrain coupled with the shallow grazing angle results in a variable resolution, occluded, sparse range data that is hard to register without some a-priori knowledge of the environment. 

\begin{figure}[htb!]
\centering
    \subfloat[\label{subfig-1:dummy}]{%
    \begin{minipage}[c][0.8\width]{0.49\textwidth}
    	\centering 
      \includegraphics[width=1.0\textwidth]{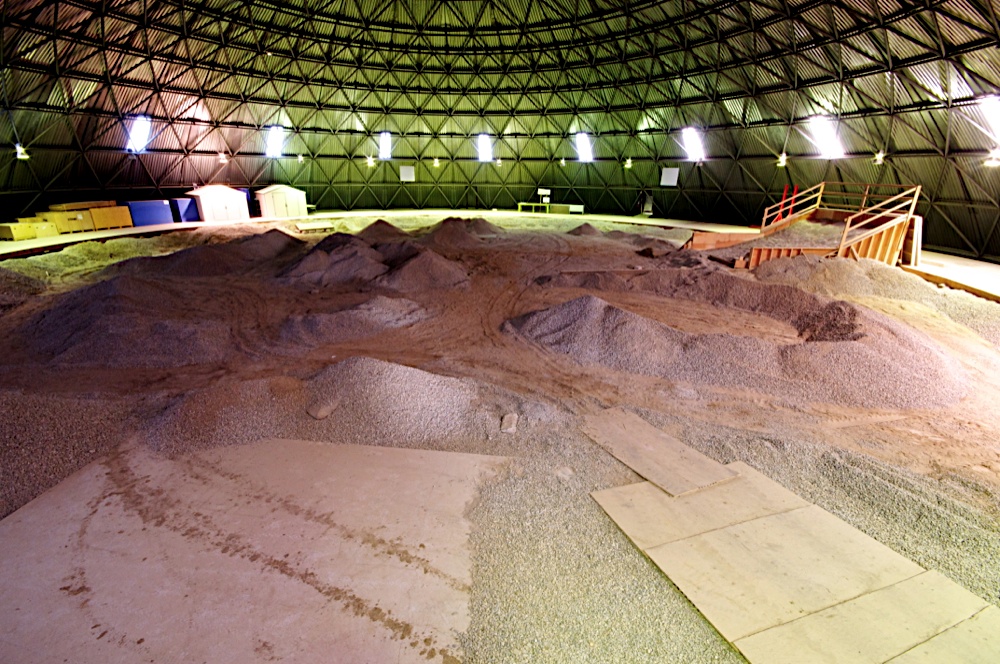}
      \end{minipage}
    }
    \subfloat[\label{subfig-2:dummy}]{%
    	\begin{minipage}[c][0.8\width]{0.49\textwidth}
    	\centering
      \includegraphics[width=0.8\textwidth]{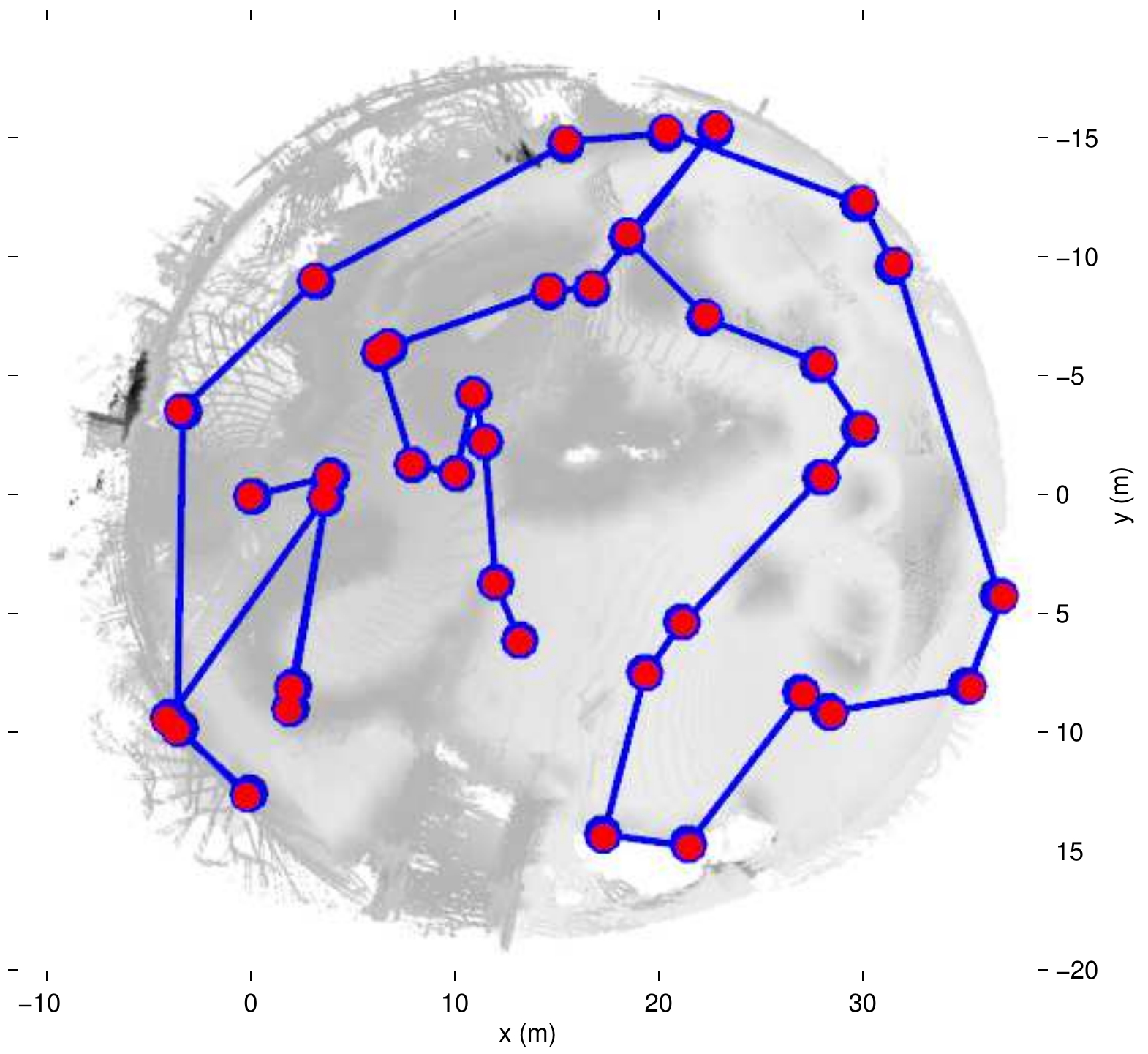}
      \end{minipage}
    }

    \caption{(a) UTIAS Mars-dome indoor planetary analogue environment (Photograph courtesy of Dr. Tim Barfoot, Autonomous Space Robotics Lab, University of Toronto), and (b)  Overhead view of the UTIAS Mars-dome terrain and the rover scan locations. 
    }
    \label{fig:utias_a100_dome_environment}
\end{figure}

The University of Toronto Institute for Aerospace Studies (UTIAS) Mars-dome analogue indoor data-set (the a100\_dome) \cite{2011-chi-dataset} consists of 95 scans obtained by vertically scanning SICK LMS291-S05 laser rangefinder with a vertical resolution of 0.5 degrees, at the rover test facility at Toronto, Ontario. The environment within the dome emulates unstructured Mars-like terrain with sand, gravel, and hills (see Figure \ref{fig:utias_a100_dome_environment}). Ground truth data is provided from four retro-reflective markers. Every 5th pair of scan was used for registration with an average distance traveled between consecutive scans of 2.38 meters, and an average rotation of 0.45 radians.

\begin{figure}[htb!]
\centering
    \subfloat[\label{subfig-1:dummy}]{%
    	\begin{minipage}[c][0.5\width]{0.5\textwidth}
    	\centering    	
      \includegraphics[width=1\textwidth]{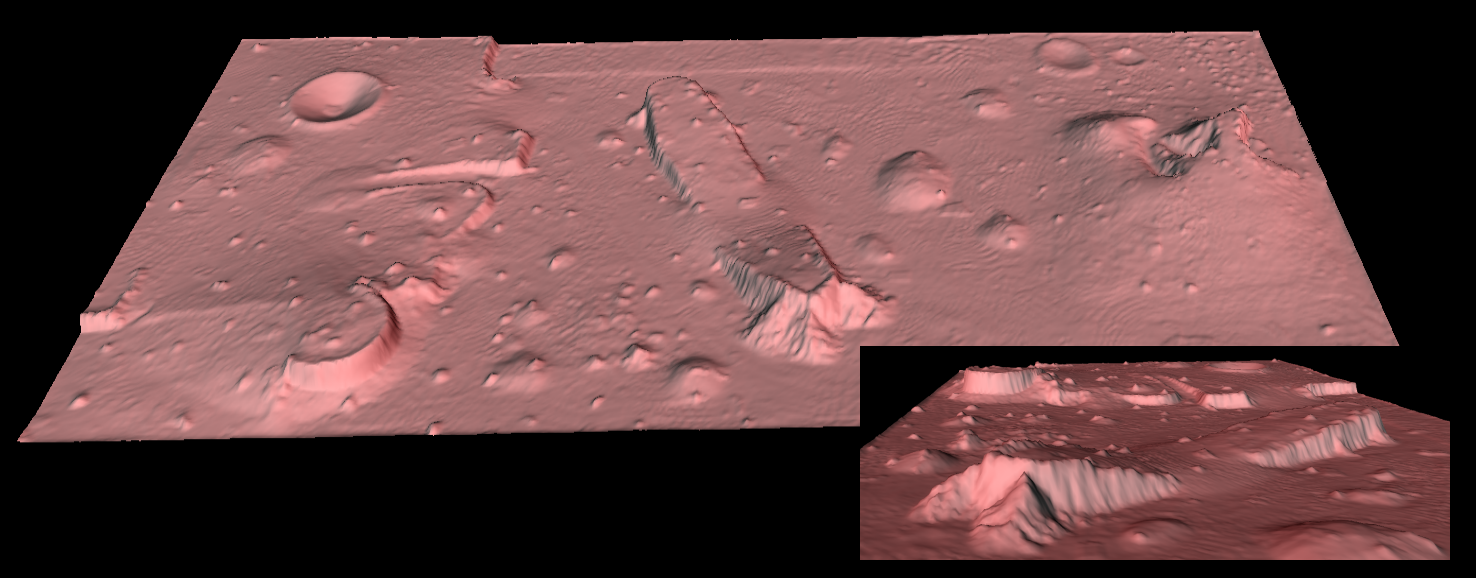}
      \end{minipage}
    }
    \subfloat[\label{subfig-2:dummy}]{%
    	\begin{minipage}[c][0.5\width]{0.5\textwidth}
    	\centering
      \includegraphics[width=1\textwidth]{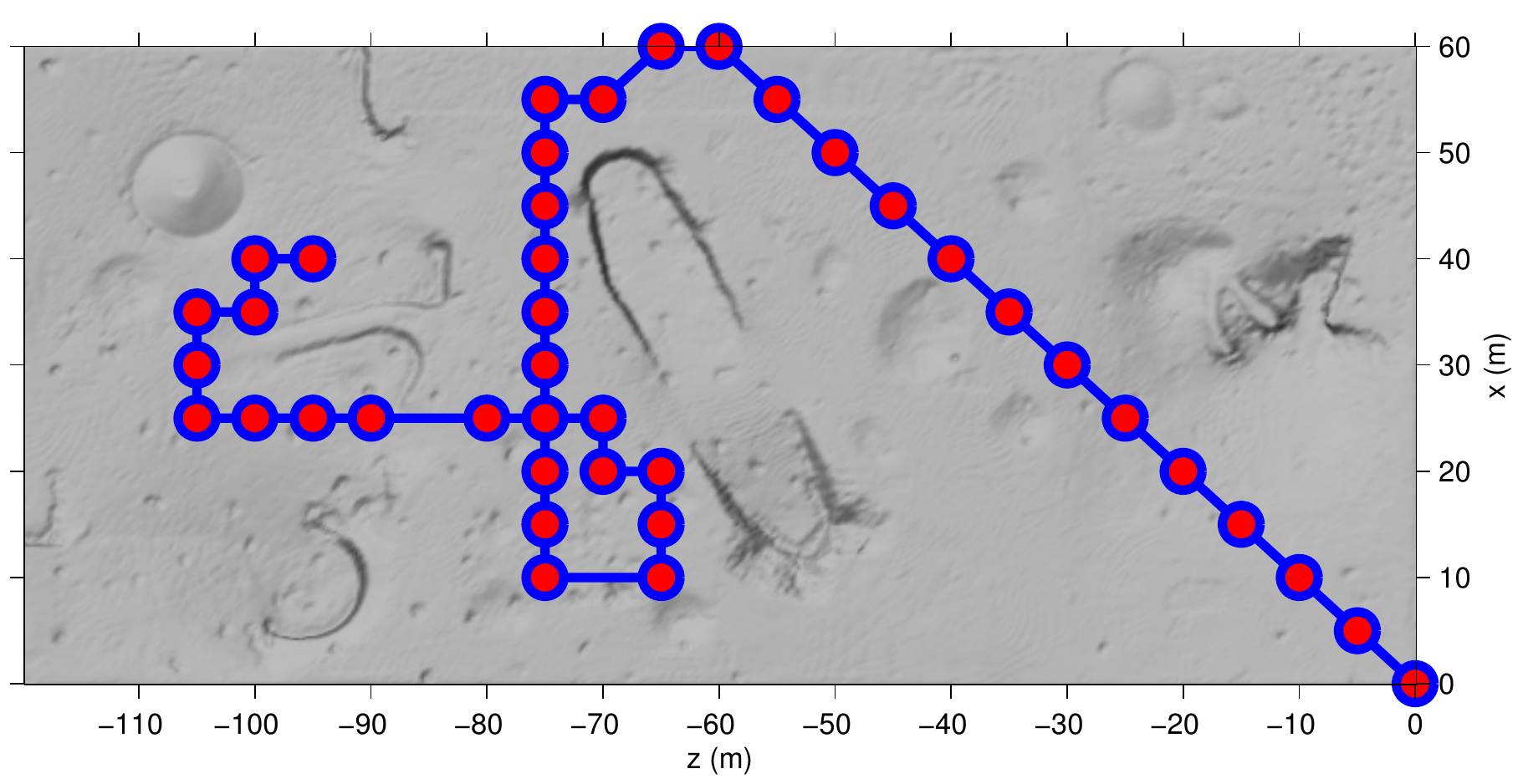}
      \end{minipage}
    }
    \caption{(a) Simulated CSA Mars Yard environment, and (b)  Overhead view of the CSA Mars emulation terrain and the rover scan locations. }
    \label{fig:ivigms_sim_environment}
  \end{figure}

Neptec Design Group has recently developed the IVIGMS 3D LIDAR sensor for the Canadian Space Agency, which can be reconfigured on the fly to produce both long range sparse point clouds and short range high density clouds with the same sensor. This unique flexibility makes the IVIGMS LIDAR an interesting option for planetary navigation. The IVIGMS 3D laser scanner is tuned with a set of pre-programmed beam trajectories. At the start of the mission, a sparse local map is generated, and low rate sparse scan data, captured while traveling through the terrain, is used to match to the original map and track the progress. The planetary analogue outdoor simulated data-set consists of 40 scans obtained by simulating the IVIGMS laser rangefinder, as depicted in Figure~\ref{fig:ivigms_sim_environment}. The scans were taken from the simulated CSA Mars emulation terrain, for which a digital elevation map (DEM) of dimensions 60m x 120m at 25cm resolution was made available. The emulation terrain includes unstructured Mars-like surface elements constructed of sand and gravel, and containing ridges, hills and a crater.  Ground truth data consists of absolute sensor pose data provided by the IVIGMS simulator. Consecutive scan pairs were used for registration with an average linear distance traveled between consecutive scans of 5.4 meters and an average rotation of 0.09 radians.

\begin{figure}[htb!]
      \centering
      \subfloat[\label{subfig-1:dummy}]{%
      \includegraphics[width=0.49\textwidth]{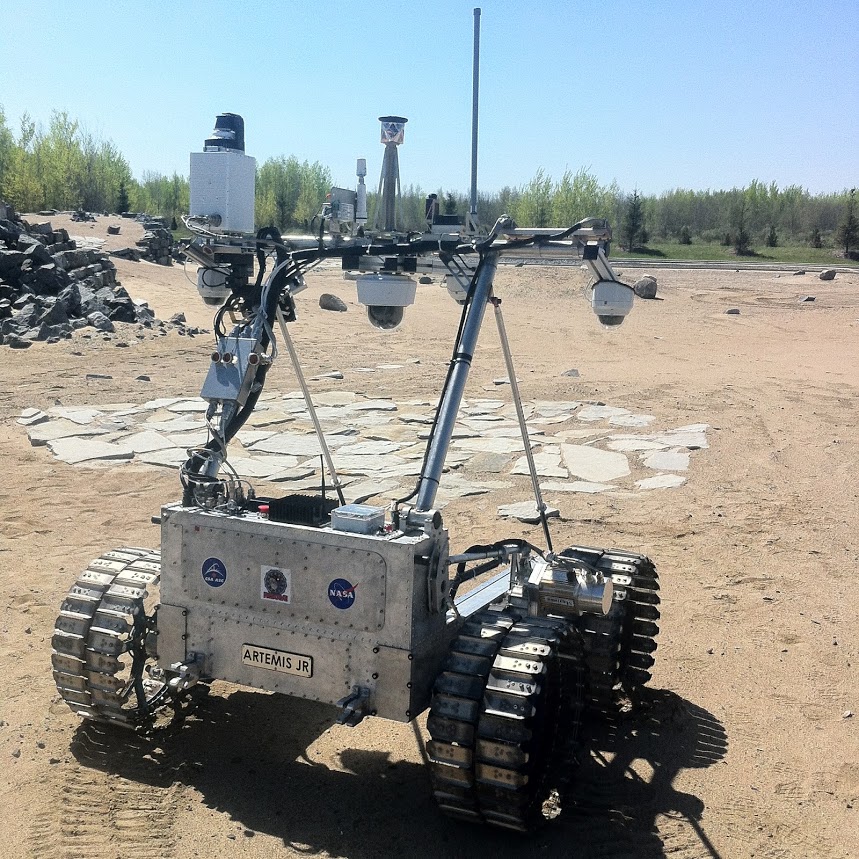}
    }
    \subfloat[\label{subfig-2:dummy}]{%
      \includegraphics[width=0.25\textwidth]{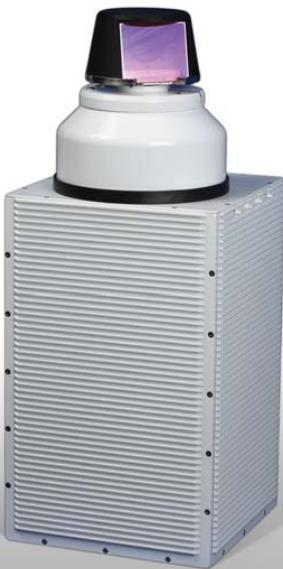}
    }
      \caption{(a) Artemis Junior robot at the CSA mars yard with Neptec IVIGMS laser range finder mounted top-left, and (b) Closeup of the IVIGMS sensor.}
      \label{artemis_junior_ncfrn}
\end{figure}

\begin{figure}[htb!]
\centering
    	\subfloat[\label{subfig-1:dummy}]{%
    \begin{minipage}[c][0.8\width]{0.49\textwidth}
    	\centering 
      \includegraphics[width=1\textwidth]{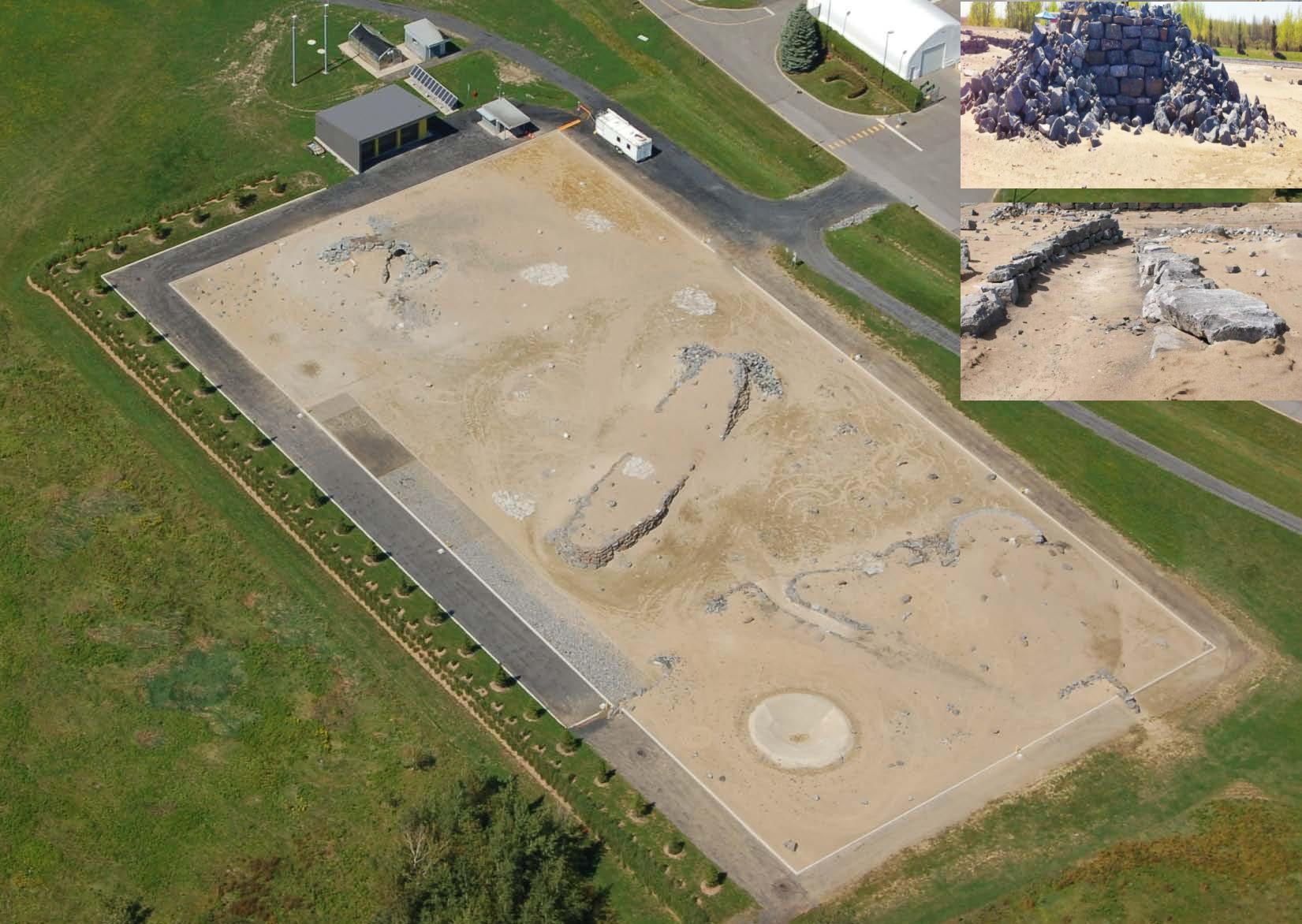}
      \end{minipage}
    }
    \subfloat[\label{subfig-2:dummy}]{%
    	\begin{minipage}[c][0.8\width]{0.5\textwidth}
    	\centering
      \includegraphics[width=1\textwidth]{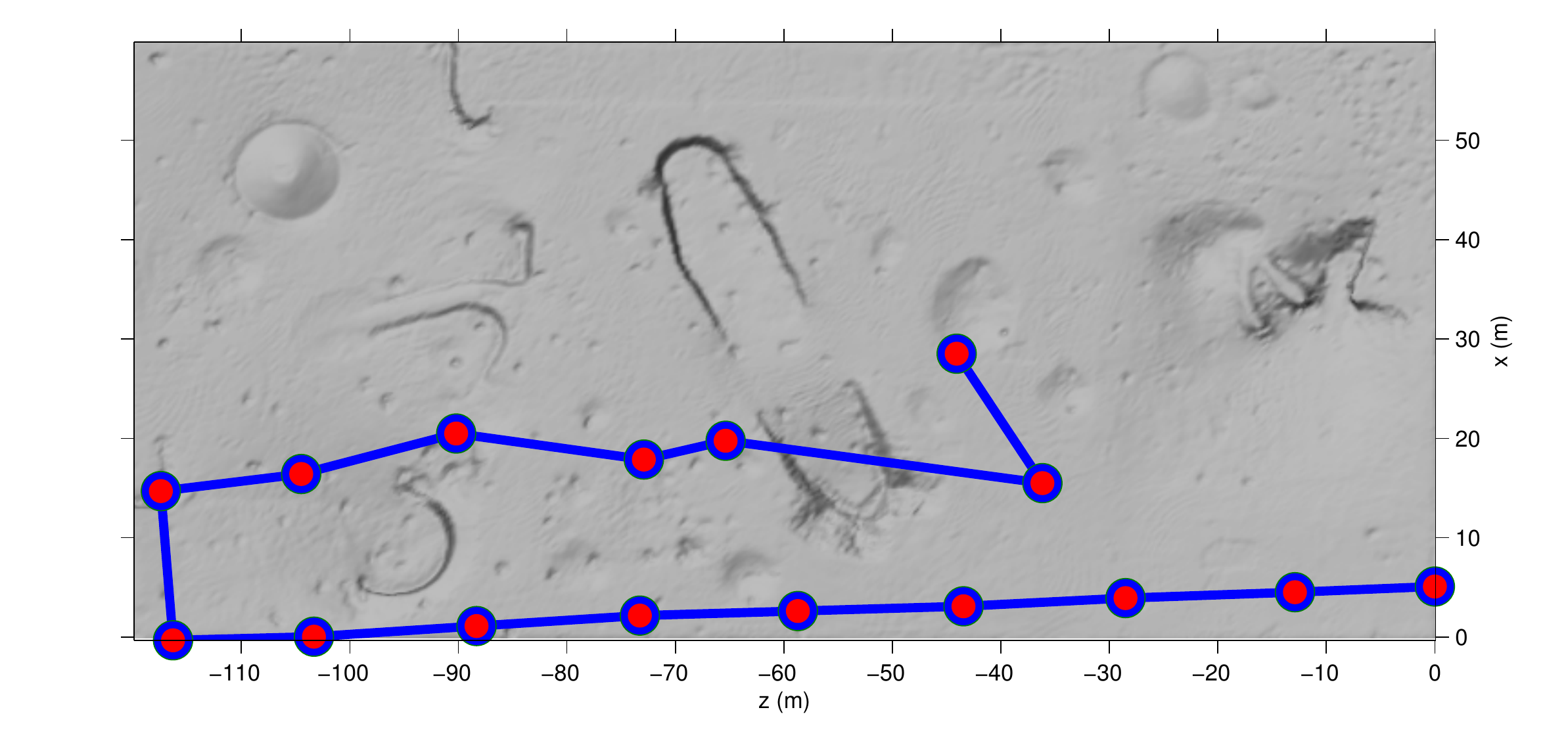}
      \end{minipage}
    }
    \caption{(a) Real world CSA Mars Yard environment (Photograph courtesy of CSA), and (b)  Overhead view of the CSA Mars emulation terrain and the rover scan locations. }
    \label{fig:ivigms_real_world_environment}
  \end{figure}

The real-world outdoor planetary analogue data-set consists of 17 scans obtained from IVIGMS laser rangefinder mounted on the Artemis Junior robot provided by CSA, as depicted in Figure~\ref{artemis_junior_ncfrn}. The scans were taken at the CSA Mars emulation terrain of dimensions 60m x 120m (See Figure \ref{fig:ivigms_real_world_environment}). Laser scan data was aggregated for 40 seconds at the rate of 25KHz with a field of view 360$\degree$ x 45$\degree$, for approximately 15 meter rover displacement in the yard. Laser scan data was later trimmed to a quarter of the number of points to get sparse scans.

In addition to the rough terrain, sparse features, large inter-scan transformation error, and occlusions, it was found that the laser scans exhibited a high degree of noise. The majority of the noise consists of outliers, bad reflections and inaccurate integration of points. The exact source of the noise is currently being investigated and a thorough analysis of measurement noise level is needed to generate a reliable noise model to aid in automated de-noising using algorithms such as RANSAC \cite{1981-ransac}. In this paper, data points corresponding to measurement errors and points lying outside the physical boundary of the Mars yard were manually removed. Ground truth data consists of absolute sensor pose data provided by manual registration of each scan.  

The first algorithm used for comparison is based on the work of Rusu et al. \cite{2008-rusu-pfh}. FPFH features are computed over the scan-pair with a search radius set to 1m, and an initial alignment is determined from the SAC-IA method. Points are sampled from each cloud with the pairwise distance greater than 2.5m. Next, ICP is used to refine the transformation based on the initial guess from SAC-IA results to obtain the final transformation estimation with the maximum correspondence distance set to 5m. 

Corner-based Harris-3D \cite{1988-harris3d} features are combined with FPFH descriptors to form the second algorithm used for comparison. The search radius used to determine the corner points in the scan-pair is set to 1m, and the k-value for the nearest neighbor search around these key-points is set to 20. Correspondences between the scans are found using nearest neighbor search. Next, correspondences are filtered using RANSAC with the maximum distance between the corresponding points (inlier threshold) set to 15m. The initial transformation is determined from these correspondences using SVD followed by the ICP refinement with maximum correspondence distance set to 5m as in the first comparative algorithm.

The selection of the parameters for both the aforementioned algorithms is a valid concern. In order to provide a fair comparison for the competing algorithms, we performed a test using all the scans of these data-sets and determined the parameter values based on the best scan registration accuracy results. For all algorithms reference implementations are provided in the Point Cloud Library (PCL) \cite{2011-rusu-pcl}. The maximum number of iterations was set to 100 and the optimization was terminated when the norm of the gradient or the norm of the step size falls below $10^{-6}$. The absolute error in rotation and translation is compared with the ground truth measurements. Despite some occlusions, there is a significant overlap  (greater than 50\%) between the scans as the sensing range exceeds the bounds of the physical environment in these data-sets.\\

The experimental results are presented in two parts. First, we evaluate the scan registration accuracy and quality of the generated maps. Second, a run-time comparison for all algorithms is presented followed by a discussion based on the results.

\subsection{Evaluation of Scan Registration Accuracy}

\begin{figure}[htb!] %H
\centering
\subfloat[\label{subfig-1:dummy}]{%
    \begin{minipage}[c][0.8\width]{0.49\textwidth}
    	\centering 
      \includegraphics[width=1\textwidth]{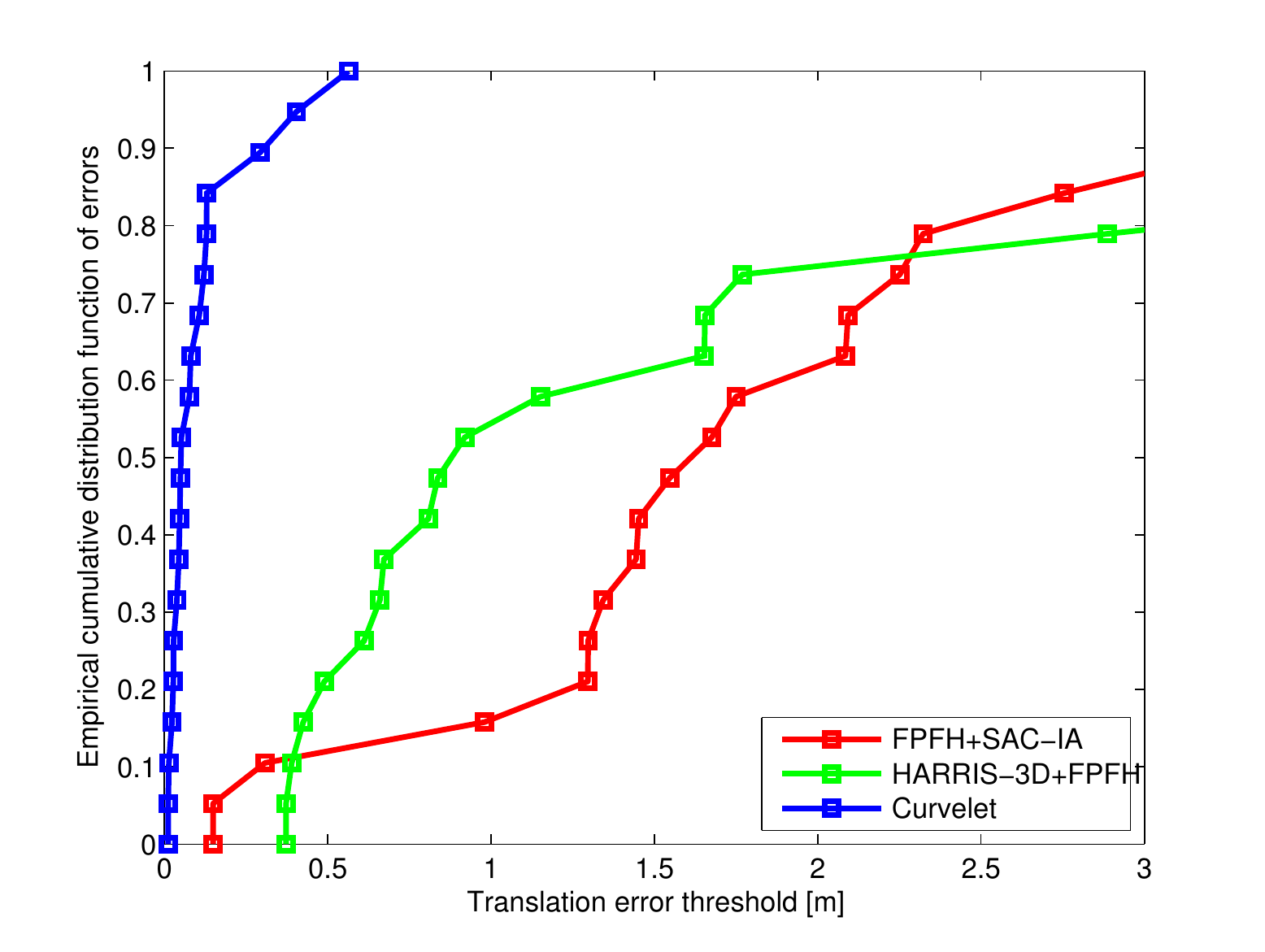}
      \end{minipage}
    }
    \subfloat[\label{subfig-2:dummy}]{%
    	\begin{minipage}[c][0.8\width]{0.5\textwidth}
    	\centering
      \includegraphics[width=1\textwidth]{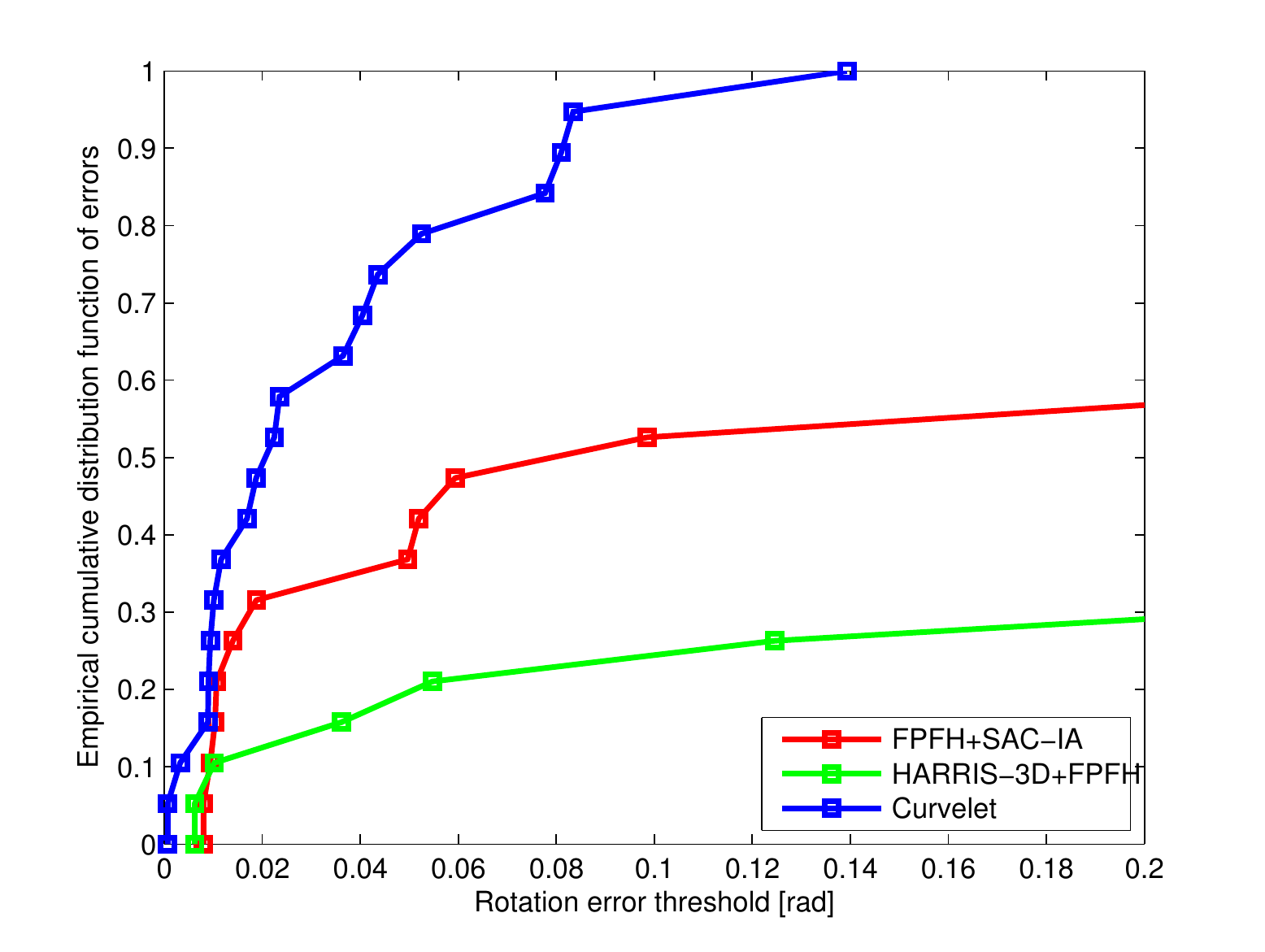}
      \end{minipage}
    }
    \caption{Error distributions for UTIAS Mars-Dome indoor data-set - a100\_dome. (a) Translation error (m). (b) Rotational error (rad).}
    \label{fig:utias_a100_dome_ecdf}
  \end{figure}

\begin{figure}[htb!]
\centering
\subfloat[\label{subfig-1:dummy}]{%
    \begin{minipage}[c][0.8\width]{0.49\textwidth}
    	\centering 
      \includegraphics[width=1\textwidth]{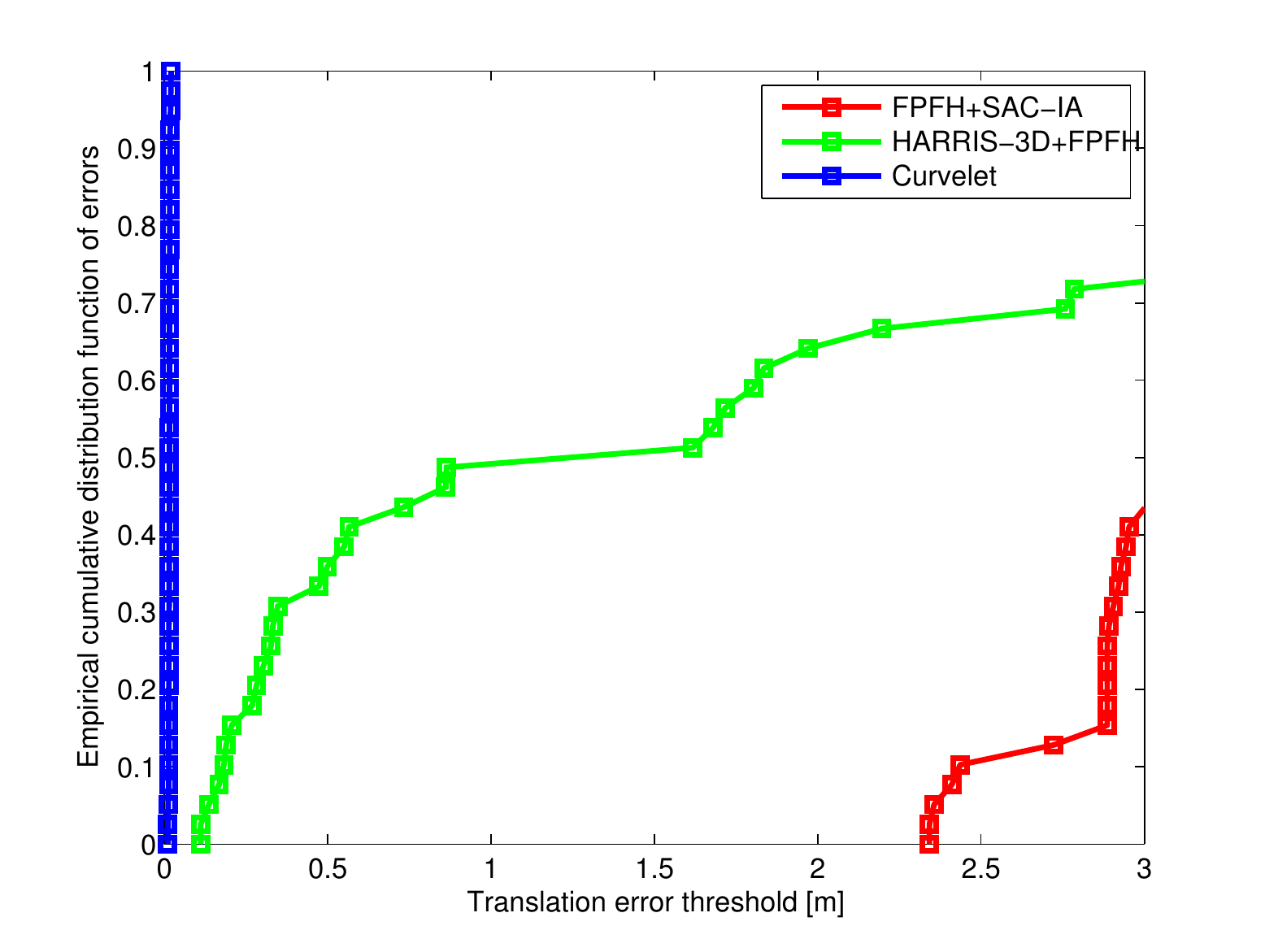}
      \end{minipage}
    }
    \subfloat[\label{subfig-2:dummy}]{%
    	\begin{minipage}[c][0.8\width]{0.5\textwidth}
    	\centering
      \includegraphics[width=1\textwidth]{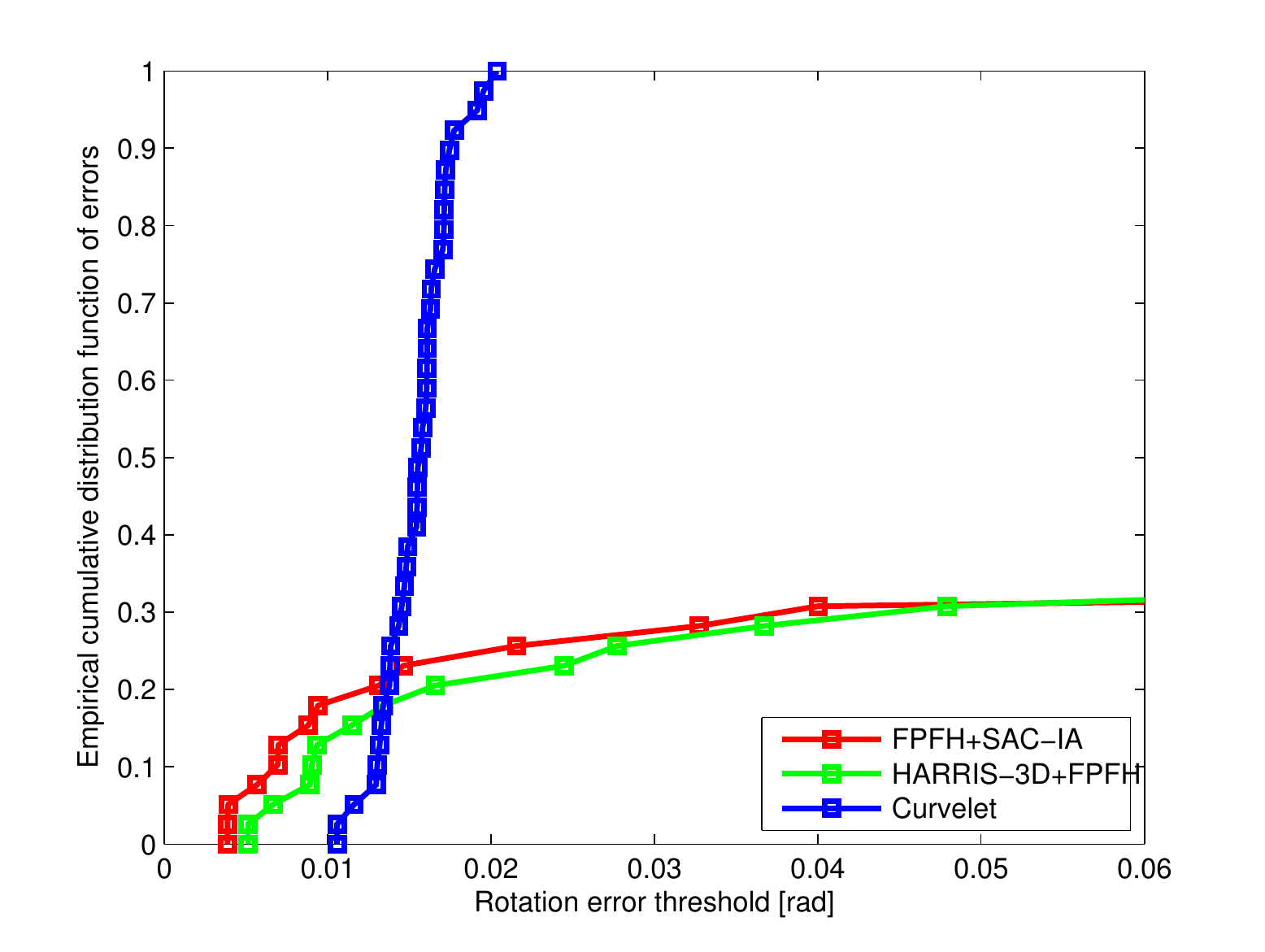}
      \end{minipage}
    }
    \caption{Error distributions for IVIGMS simulated CSA Mars-Yard outdoor data-set. (a) Translation error (m). (b) Rotational error (rad).}
    \label{fig:ivigms_simulated_ecdf}
  \end{figure}

\begin{figure}[htb!]
\centering
\subfloat[\label{subfig-1:dummy}]{%
    \begin{minipage}[c][0.8\width]{0.49\textwidth}
    	\centering 
      \includegraphics[width=1\textwidth]{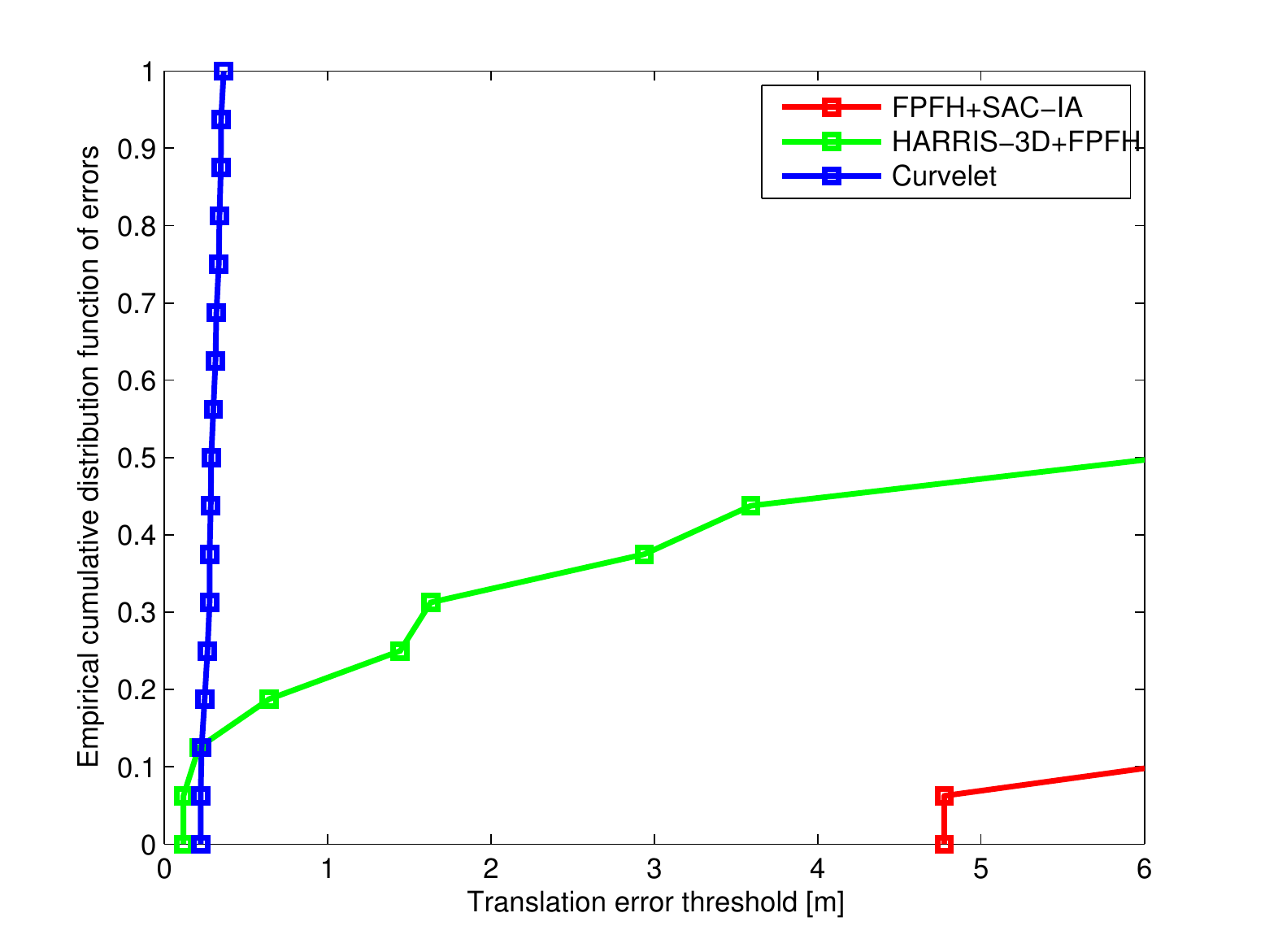}
      \end{minipage}
    }
    \subfloat[\label{subfig-2:dummy}]{%
    	\begin{minipage}[c][0.8\width]{0.5\textwidth}
    	\centering
      \includegraphics[width=1\textwidth]{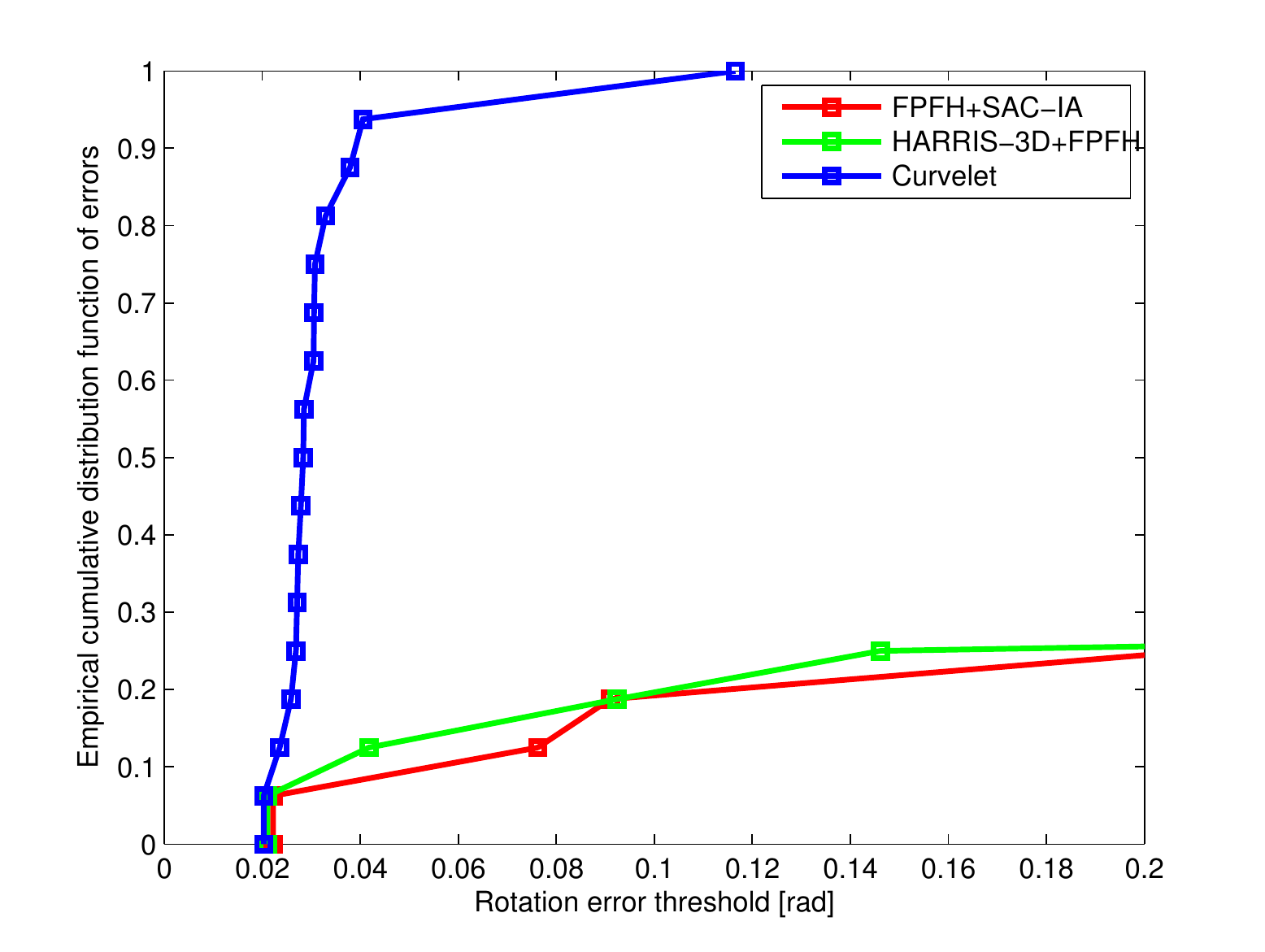}
      \end{minipage}
    }    
    \caption{Error distributions for IVIGMS real-world CSA Mars-Yard outdoor data-set. (a) Translational error (m). (b) Rotational error (rad).}
    \label{fig:ivigms_real_world_ecdf}
  \end{figure}

In order to evaluate the accuracy of the scan registration, root mean squared error (RMSE) is computed for both translation and rotation with respect to the ground truth. Assuming that the error samples are independent and identically-distributed random variables, the asymptotic behavior of the empirical measure of the error samples can be determined using the empirical cumulative distribution function \cite{gilvenko-cantelli-1959}. It is defined as the proportion of error samples less than or equal to a given error metric, converging to a cumulative probability of 1 as the error magnitude increases. Intuitively it can be viewed as the proportion of successful scan registrations below the error threshold. Figures \ref{fig:utias_a100_dome_ecdf}, \ref{fig:ivigms_simulated_ecdf}, and \ref{fig:ivigms_real_world_ecdf} show the empirical cumulative distribution function of error samples (y-axis) for the given registration error threshold (along x-axis). The error distributions demonstrate that curvelet transform based scan registration converges faster to a maximum probability of one, and produces the most consistently accurate results when compared to FPFH+SAC-IA and Harris-3D+FPFH methods. In fact, from a registration perspective, errors in rotation in excess of 0.1 radians (5.73 degrees) can be considered registration failures. Although some failures can be detected and/or rejected by robust back-end loop closure techniques such as~\cite{grisetti2012robust}, it is preferable to avoid such failures in scan-to-scan registration in the first place. 
The root mean squared errors (RMSE) in translation and rotation for the three data-sets are given in Tables \ref{table:utias_a100_dome_errors}, \ref{table:ivigms_sim_errors} and \ref{table:ivigms_real_world_errors}.
The large translational and rotational error in FPFH+SAC-IA can be attributed to the fact that FPFH is not view-point invariant. The quality of alignment is directly dependent on the point selection strategy and accurate normals computation. Both FPFH+SAC-IA and Harris-3D+FPFH algorithms show high translational and rotational errors due to the lack of feature-rich regions in the scans and absence of planar structures. Compared with other algorithms, curvelet transform based scan registration has a lower RMSE in both translation and rotation. 

\begin{figure}[htb!]
\centering
\subfloat[\label{subfig-1:dummy}]{%
    \begin{minipage}[c][0.8\width]{0.49\textwidth}
    	\centering 
      \includegraphics[width=1\textwidth]{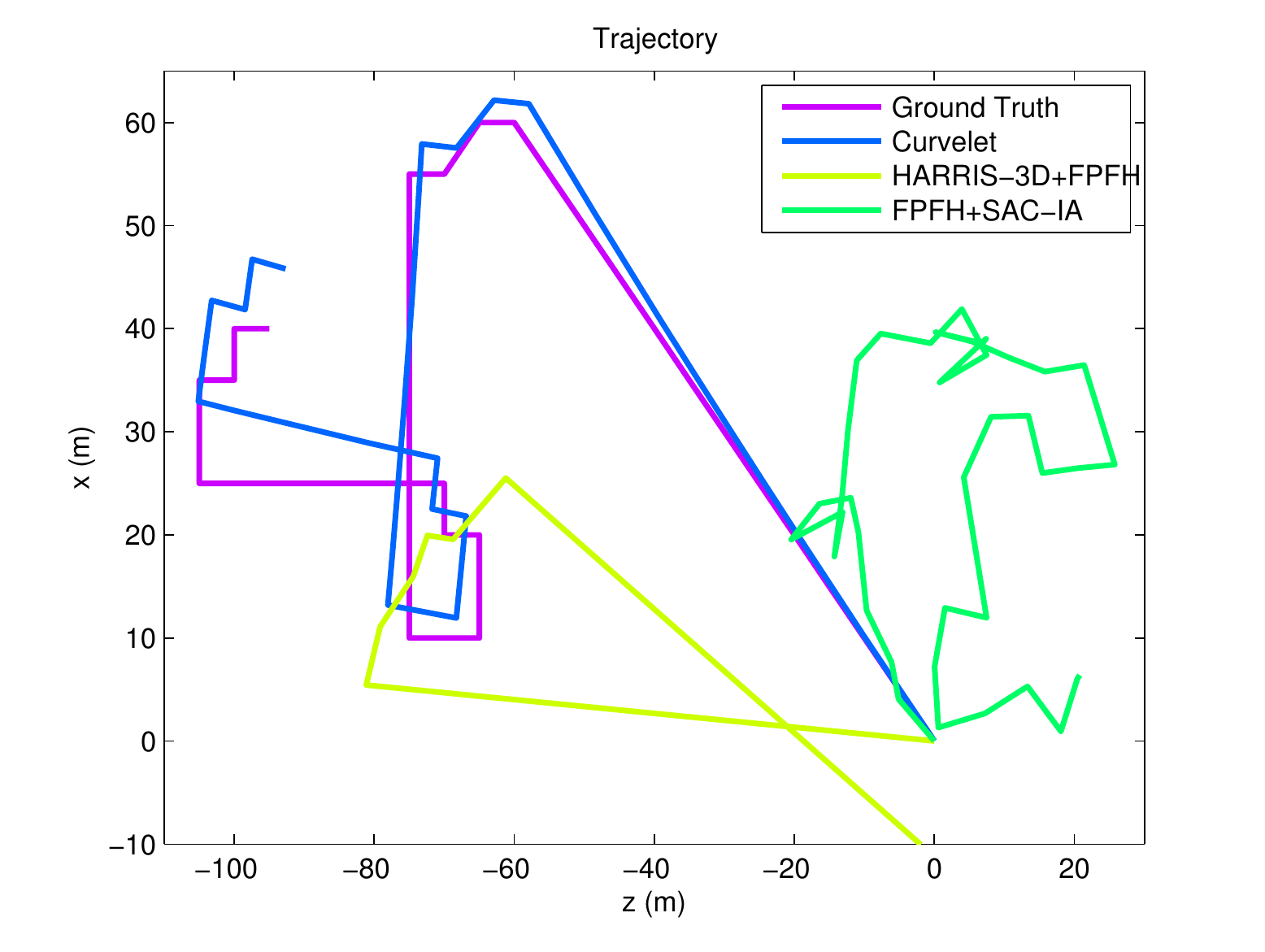}
      \end{minipage}
    }
    \subfloat[\label{subfig-2:dummy}]{%
    	\begin{minipage}[c][0.8\width]{0.5\textwidth}
    	\centering
      \includegraphics[width=1\textwidth]{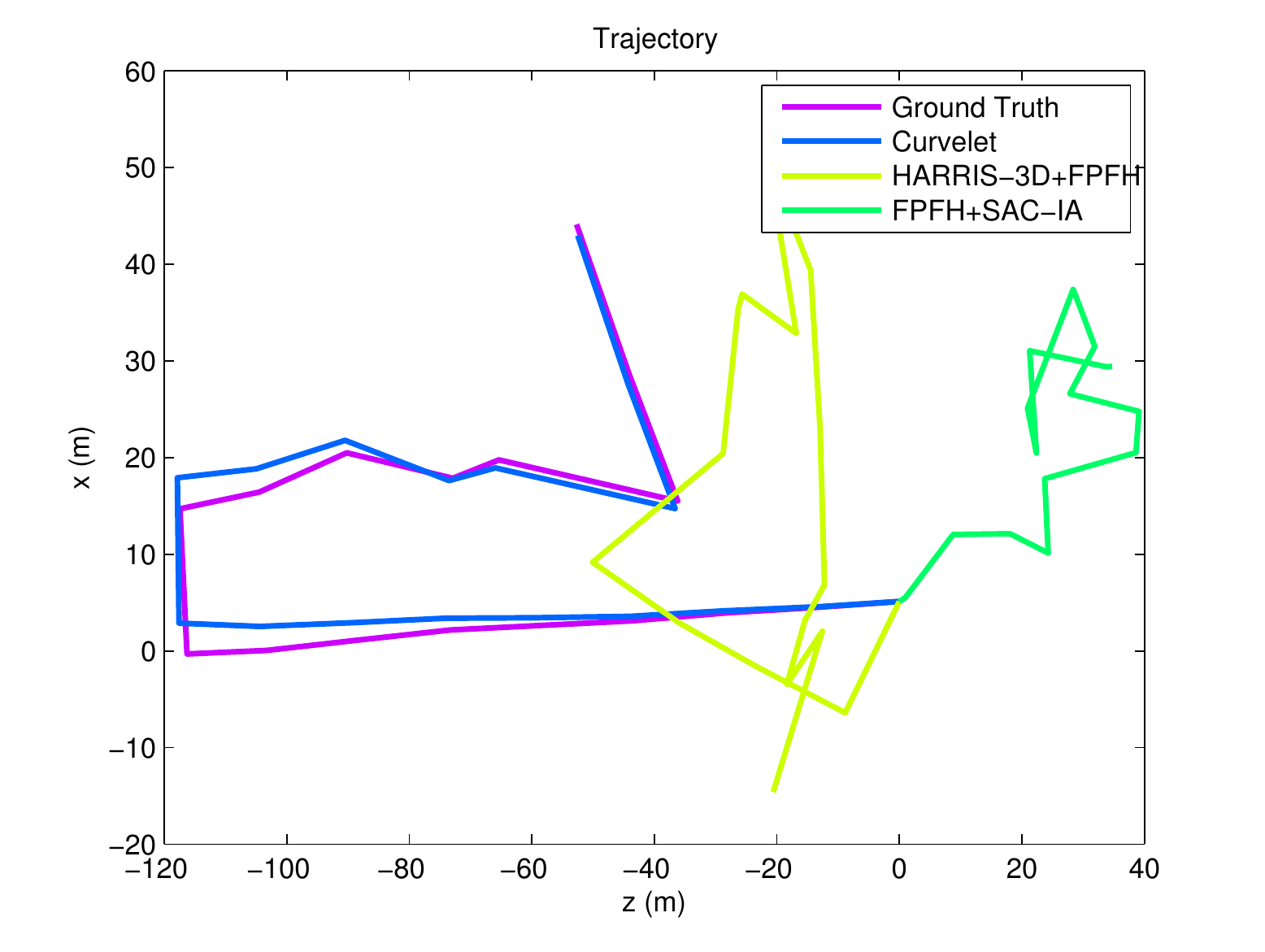}
      \end{minipage}
    }
    \caption{Visualization of the rover trajectory. (a) IVIGMS simulated CSA Mars-Yard outdoor data-set. (b) IVIGMS real-world CSA Mars-Yard outdoor data-set.}
    \label{fig:trajectory} 
  \end{figure}

\begin{figure}[!htb]
\centering
\subfloat[\label{subfig-1:dummy}]{%
    \begin{minipage}[c][0.8\width]{0.49\textwidth}
    	\centering 
      \includegraphics[width=1\textwidth]{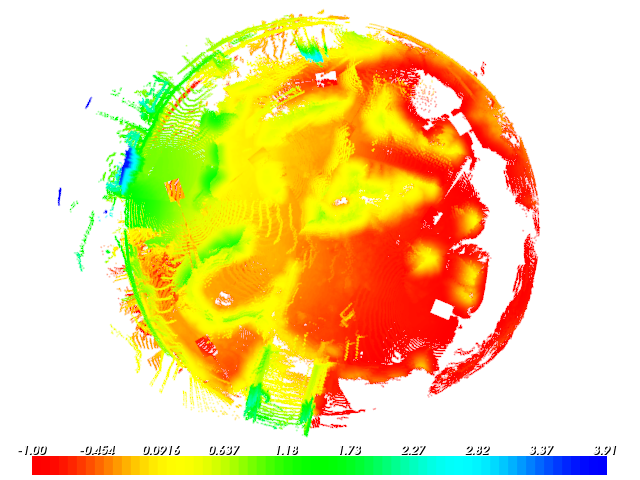}
      \end{minipage}
    }
    \subfloat[\label{subfig-2:dummy}]{%
    	\begin{minipage}[c][0.8\width]{0.5\textwidth}
    	\centering
      \includegraphics[width=1\textwidth]{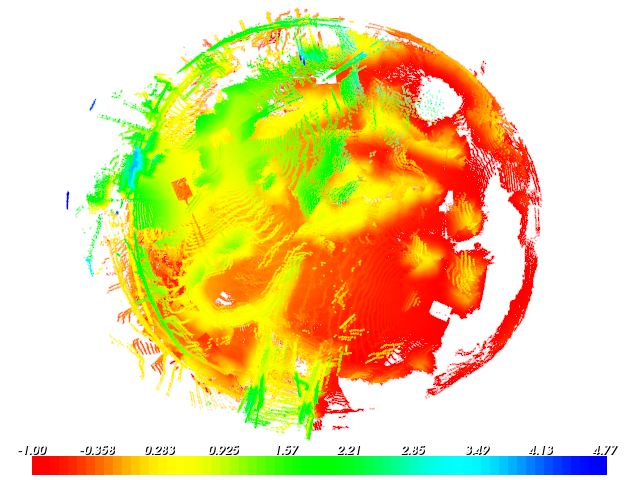}
      \end{minipage}
    }
    \caption{Final maps generated by integrating registered scans for the UTIAS Mars-Dome indoor data-set - a100\_dome. Points above -1 meter in z-height are displayed for better visualization (intensity scaled in z-axis from low-red to high-blue). (a) Ground truth.  (b) Proposed method. %(c) G-ICP map. (d) NDT map.
    }
    \label{fig:utias_a100_dome_maps}
  \end{figure}

\begin{figure}[!htb]
\centering
\subfloat[\label{subfig-1:dummy}]{%
    \begin{minipage}[c][0.6\width]{0.48\textwidth}
    	\centering 
      \includegraphics[width=1\textwidth]{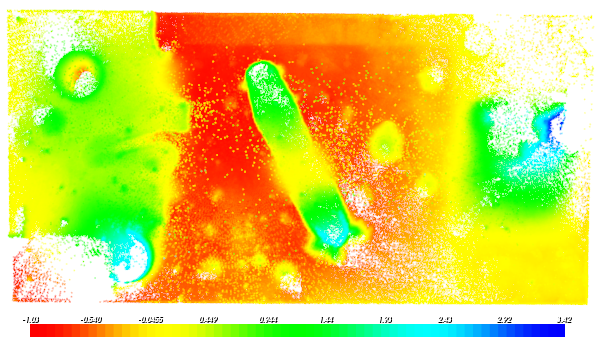}
      \end{minipage}
    }
    \subfloat[\label{subfig-2:dummy}]{%
    	\begin{minipage}[c][0.6\width]{0.48\textwidth}
    	\centering
      \includegraphics[width=1\textwidth]{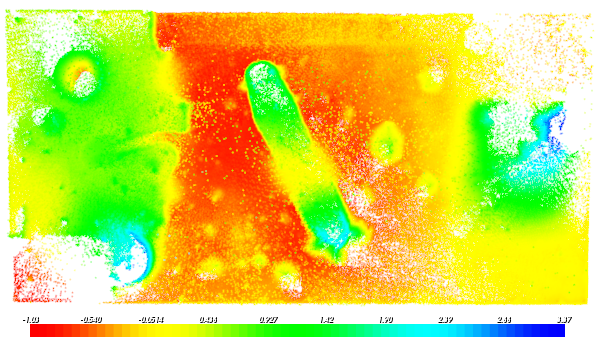}
      \end{minipage}
    }
    \caption{Final maps generated by integrating registered scans for the IVIGMS simulated CSA Mars-Yard outdoor data-set (intensity scaled in z-axis from low-red to high-blue). (a) Ground truth.  (b) Proposed method. %(c) G-ICP map. (d) NDT map (Only registered scans [0-5] are plotted to demonstrate the effect of errors in translation and rotation in building a dense map) .
    }
    \label{fig:ivigms_sim_maps}
  \end{figure}

\begin{figure}[htb!]
\centering
\subfloat[\label{subfig-1:dummy}]{%
    \begin{minipage}[c][0.6\width]{0.48\textwidth}
    	\centering 
      \includegraphics[width=1\textwidth]{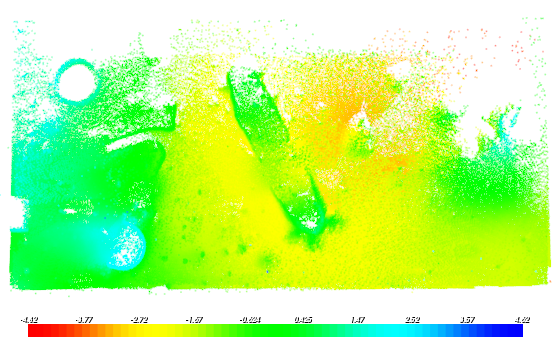}
      \end{minipage}
    }
    \subfloat[\label{subfig-2:dummy}]{%
    	\begin{minipage}[c][0.6\width]{0.48\textwidth}
    	\centering
      \includegraphics[width=1\textwidth]{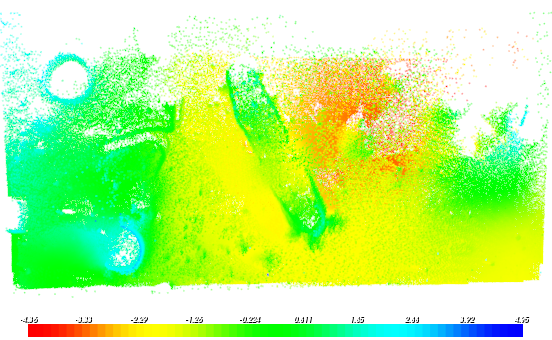}
      \end{minipage}
    }
    \caption{Final maps generated by integrating registered scans for the ground truth (intensity scaled in z-axis from low-red to high-blue) for the IVIGMS real-world CSA Mars-Yard outdoor data-set. (a) Ground truth.  (b) Proposed method. %(c) G-ICP map (Only registered scans [0-5] are plotted to demonstrate the effect of errors in translation and rotation in building a dense map). (d) NDT map (Only registered scans [0-5] are plotted to demonstrate the effect of errors in translation and rotation in building a dense map).
    }
    \label{fig:ivigms_real_world_maps}
  \end{figure}

Figure \ref{fig:trajectory} shows the rover trajectory generated by integrating the estimated rover positions from scan registration over time for the outdoor data-sets. The proposed method slowly drift away from the ground truth for the simulated outdoor data-set shown in Figure \ref{fig:trajectory}-(a), b. Comparing the trajectories between the two outdoor data-sets, it can be seen that FPFH+SAC-IA and Harris-3D+FPFH have a much higher drift in the real-world and simulated data-sets due to the inaccurate calculation of surface normals in the presence of noise and outliers. 
\begin{center}
	\begin{table}[htb!]
	\caption{Root mean squared error in translation and rotation for the UTIAS Mars-Dome indoor data-set - a100\_dome data-set.}
	\newcolumntype{L}{>{\centering\arraybackslash}m{2.00cm}}
	\newcolumntype{B}{>{\centering\arraybackslash}m{3.75cm}}
	\newcolumntype{C}{>{\centering\arraybackslash}m{5.65cm}}
	\renewcommand{\arraystretch}{1.2}% Wider
    \begin{tabular}{ |B|C|C| }
  	\hline
  	 & \textbf{RMSE Translation (m)} & \textbf{RMSE Rotation (rad)}   \\
  	 \hline
  	FPFH+SAC-IA & 1.6763  &  0.0986   \\
  	\hline
  	Harris-3D+FPFH & 0.9202  &  1.7908   \\
  	\hline
  	\textbf{Proposed Method} & 	\textbf{0.3211} & \textbf{0.0877}\\
  	\hline
  	\end{tabular}
  	\label{table:utias_a100_dome_errors}
    \end{table}
\end{center}
\begin{center}
	\begin{table}[htb!]
	\caption{Root mean squared error in translation and rotation for the IVIGMS simulated CSA Mars-Yard outdoor data-set.}
	\newcolumntype{L}{>{\centering\arraybackslash}m{2.00cm}}
	\newcolumntype{B}{>{\centering\arraybackslash}m{3.75cm}}
	\newcolumntype{C}{>{\centering\arraybackslash}m{5.65cm}}
	\renewcommand{\arraystretch}{1.2}% Wider
    \begin{tabular}{ |B|C|C| }
  	\hline
  	 & \textbf{RMSE Translation (m)} & \textbf{RMSE Rotation (rad)} \\
  	\hline
  	FPFH+SAC-IA & 3.1916  & 2.4717    \\
  	\hline
  	Harris-3D+FPFH &  1.6178 & 2.1043    \\
  	\hline
  	\textbf{Proposed Method} & 	\textbf{0.0147} & \textbf{0.0090}\\
  	\hline
  	\end{tabular}
  	\label{table:ivigms_sim_errors}
    \end{table}
\end{center}
\begin{center}
	\begin{table}[htb!]
	\caption{Root mean squared error in translation and rotation for the IVIGMS real-world CSA Mars-Yard outdoor data-set.}
	\newcolumntype{L}{>{\centering\arraybackslash}m{2.00cm}}
	\newcolumntype{B}{>{\centering\arraybackslash}m{3.75cm}}
	\newcolumntype{C}{>{\centering\arraybackslash}m{5.65cm}}
	\renewcommand{\arraystretch}{1.2}% Wider
    \begin{tabular}{ |B|C|C| }
  	\hline
  	 & \textbf{RMSE Translation (m)} & \textbf{RMSE Rotation (rad)} \\
  	\hline
  	FPFH+SAC-IA & 8.6808  & 2.3560    \\
  	\hline
  	Harris-3D+FPFH & 6.4692  & 2.5836   \\
  	\hline
  	\textbf{Proposed Method} & \textbf{0.1855} & \textbf{0.0293}\\
  	\hline
  	\end{tabular}
  	\label{table:ivigms_real_world_errors}
    \end{table}
\end{center}
Figures \ref{fig:utias_a100_dome_maps}, \ref{fig:ivigms_sim_maps}, and \ref{fig:ivigms_real_world_maps} show the final dense map generated by integrating all registered scans for the ground truth and the proposed method. It can be seen that the proposed method produces an accurate dense map of the indoor Mars-dome and outdoor CSA Mars emulation terrain. In particular, the high rotational alignment accuracy of the proposed method is visible at the center of the map in Figure \ref{fig:ivigms_sim_maps}-(b) and right edge of the map in Figure \ref{fig:utias_a100_dome_maps}-(b). Due to the high degree of error in rotation and translation, dense maps generated by FPFH+SAC-IA and Harris-3D+FPFH are not shown. Some translational and rotational error remains in the curvelet registered map, as visible at the center of the map in \ref{fig:ivigms_real_world_maps}-(b), but this is small in comparison to the errors transformation errors in other algorithms.

\subsection{Run-time Evaluation}

\begin{figure}[htb!]
\centering
    \subfloat[\label{subfig-1:dummy}]{%
      \includegraphics[width=0.48\textwidth]{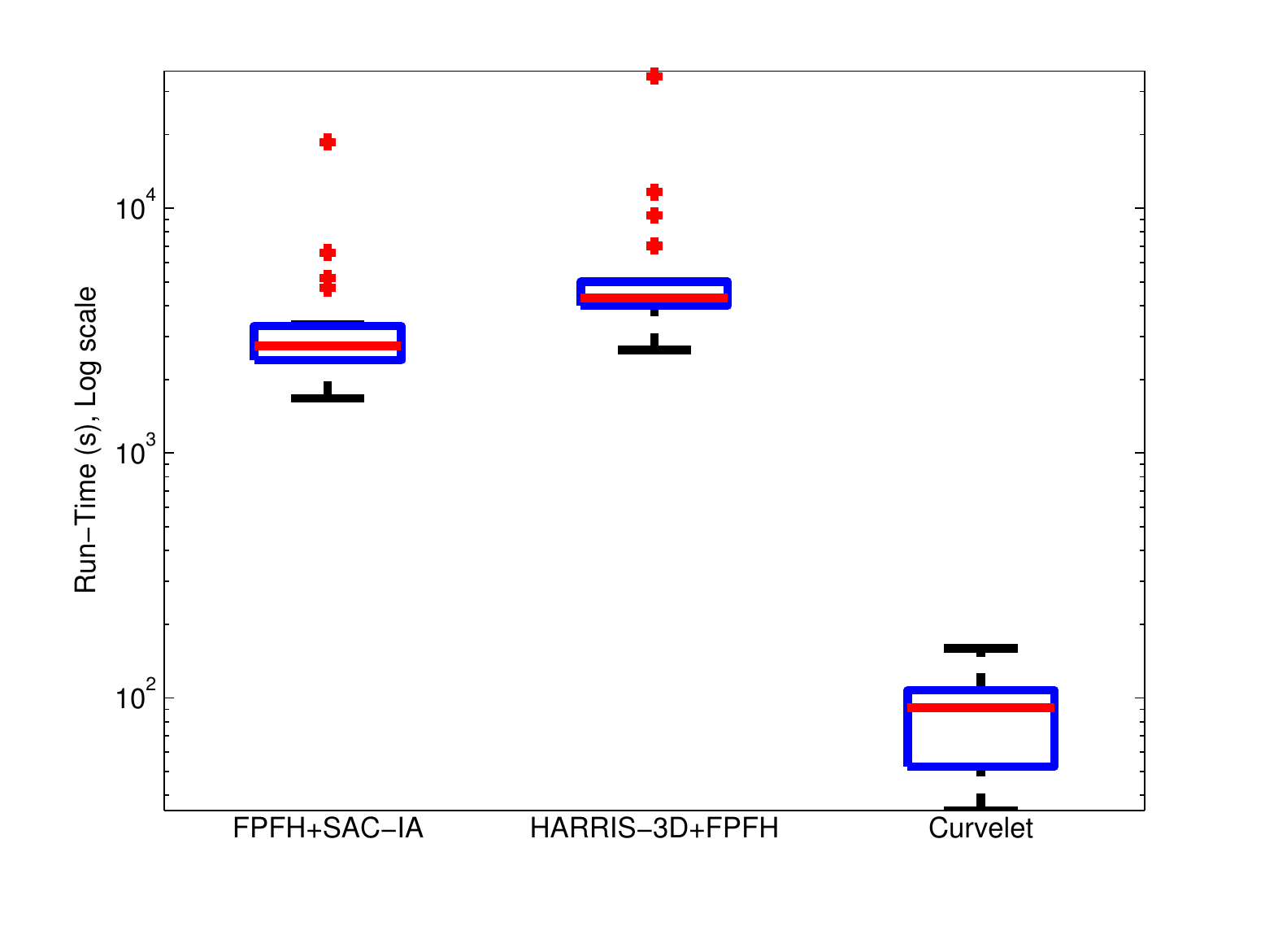}
    }
    \subfloat[\label{subfig-2:dummy}]{%
      \includegraphics[width=0.48\textwidth]{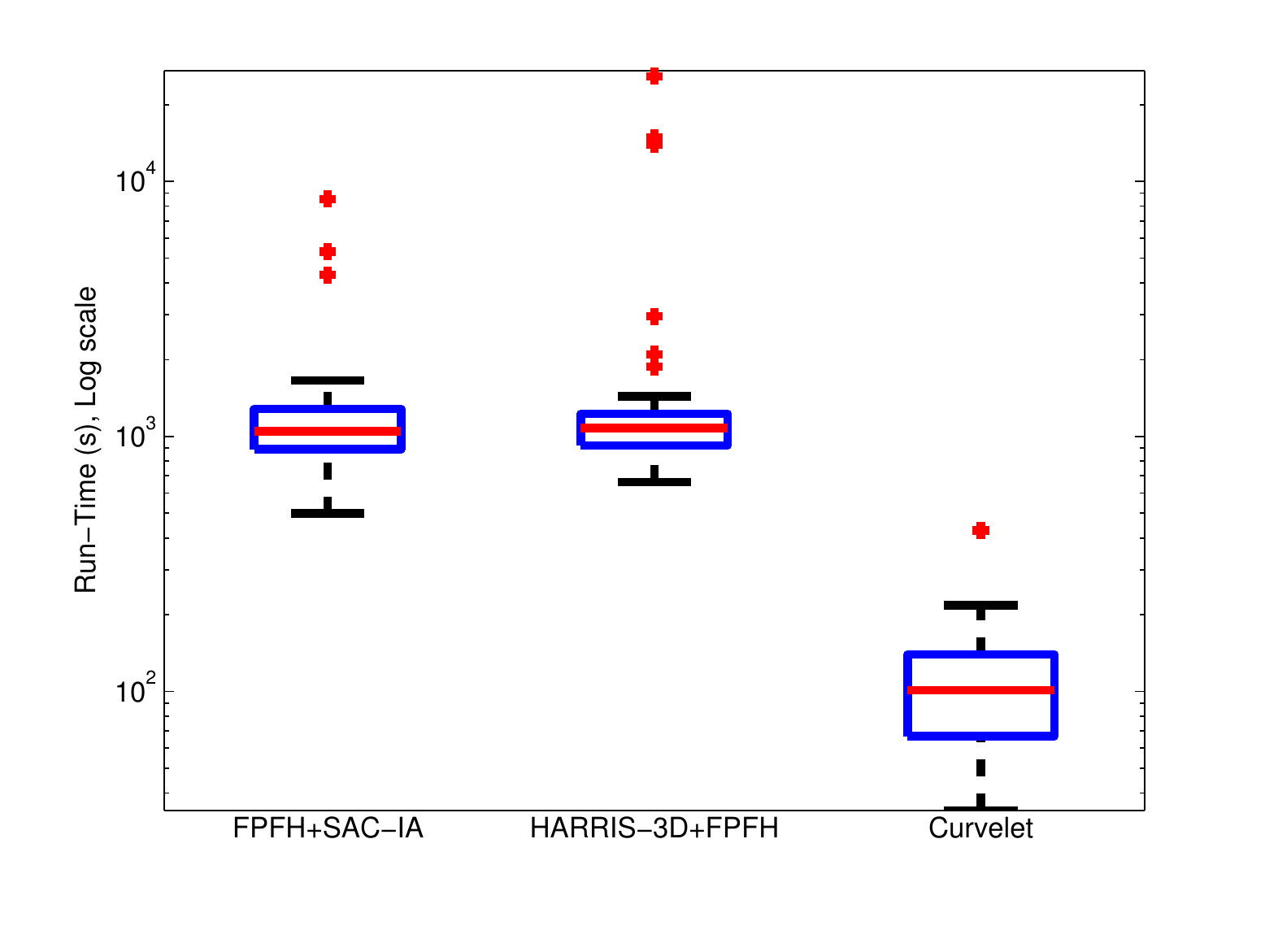}
    }
    \quad
    \subfloat[\label{subfig-3:dummy}]{%
      \includegraphics[width=0.48\textwidth]{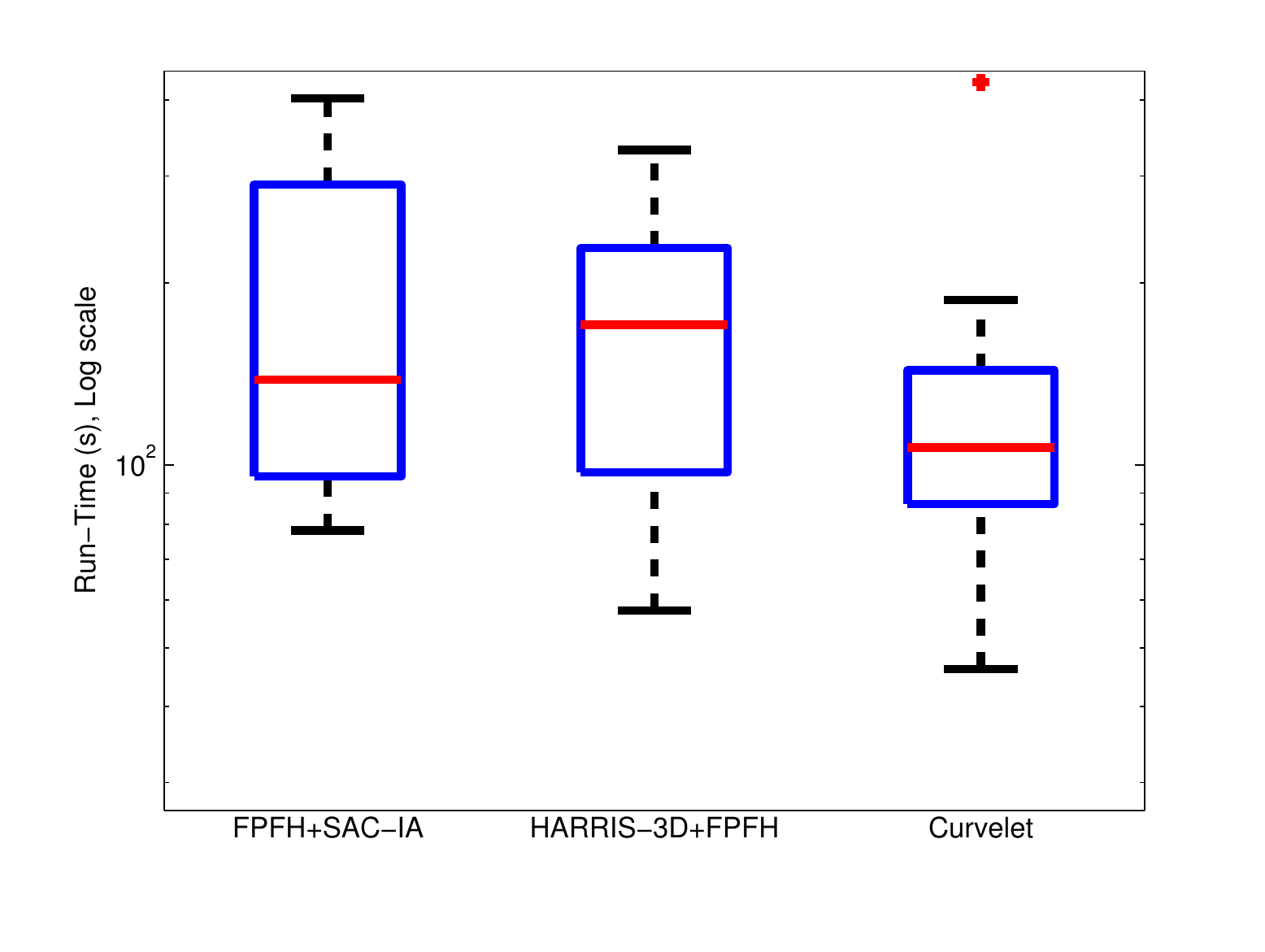}
    }
    \subfloat[\label{subfig-3:dummy}]{%
      \includegraphics[width=0.48\textwidth]{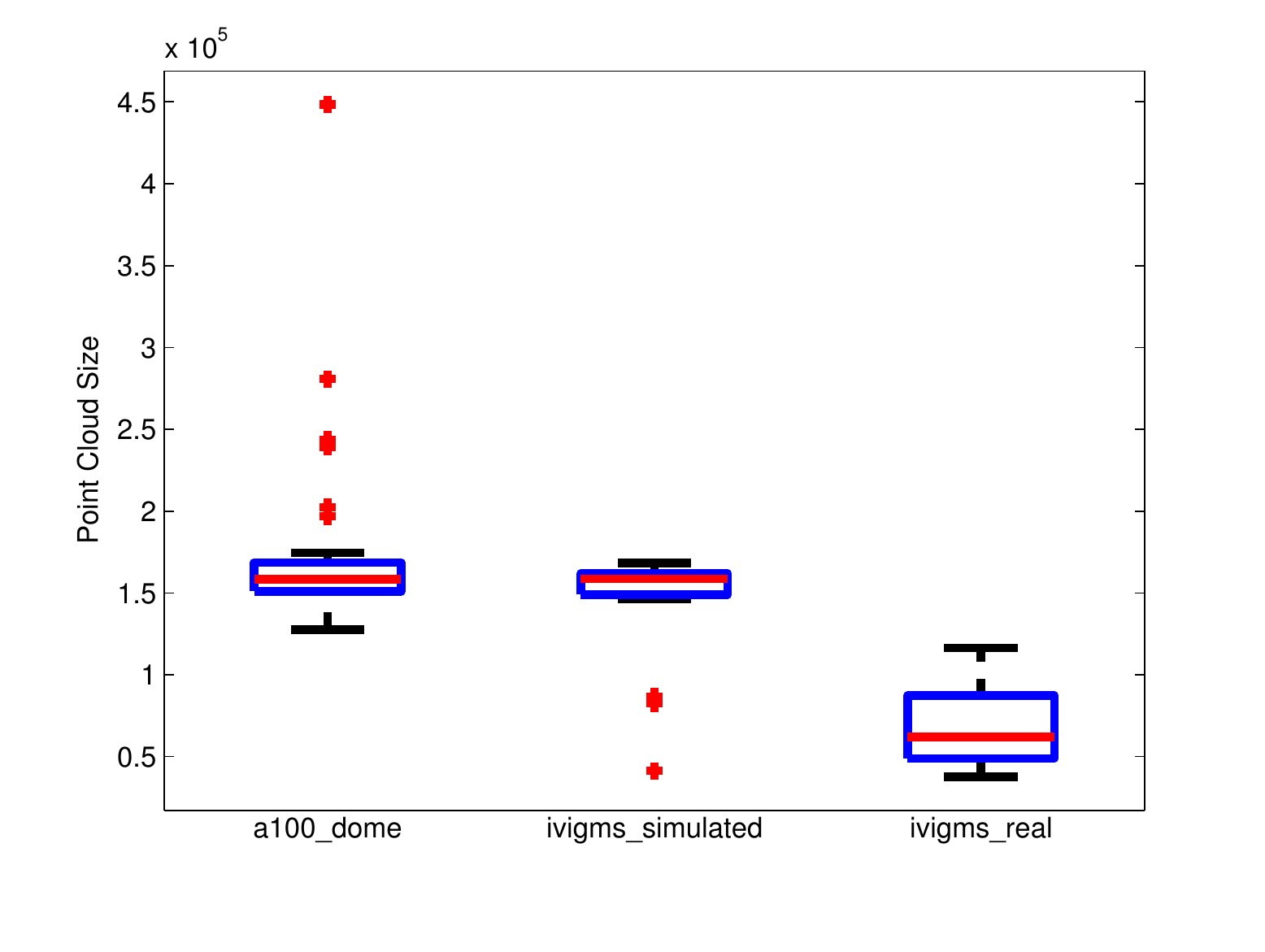}
    }
    \caption{Box-plots illustrating the scan registration run-times in log scale and the number of points in laser scans for various data-sets. (a) UTIAS Mars-Dome indoor data-set - a100\_dome. (b) IVIGMS simulated CSA Mars-Yard outdoor data-set. (c) IVIGMS real-world CSA Mars-Yard outdoor data-set. (c) Point cloud size for the three data-sets }
    \label{fig:run-time} 
  \end{figure}

Scan registration run-times for FPFH features with SAC-IA alignment, Harris-3D features with FPFH descriptors  and the proposed method for each of three data-sets is visualized in  Figure \ref{fig:run-time} as box-plots. The "central box" in box-plots represents the central 50\% of the data with a central line indicating the median, and the lower and upper boundary lines are at the first and third quartile of the data. Two vertical lines extending from the central box to 1.5 times the height of the central box indicate the remaining data not considered as outliers, and the red points on top or bottom, represent any outliers if present. The run-time evaluation clearly demonstrates that the proposed method is able to perform scan registration significantly faster than other methods. The average run-time values for each of the three data-sets are presented in Table \ref{table:run_times}. As can be seen, for all data-sets, the average run-time for the proposed method is atleast an order of magnitude faster than other methods. 

Run-times for all methods can be significantly improved by down-sampling the point cloud and changing the scale dependent parameters. However, down-sampling of the surface may result in inaccurate scan registration, and is therefore not utilized in this study. A large amount of computational time is spent in calculating normals and FPFH descriptors for FPFH+SAC-IA algorithm, and normals and key-points for the Harris-3D+FPFH algorithm. The run-time performance for both  these algorithms is directly dependent on the size of the local neighborhood selected for calculation of normals, key-points and respective descriptors. In addition parameters such as maximum correspondence distance for both SAC-IA and ICP-refinement steps, termination condition, and number of iterations are all data-set dependent. The values for these parameters mentioned above were selected based on the need to balance alignment accuracy and faster run-times. Changing these parameters may result in lower run-times at the expense of registration accuracy. The proposed method represents the 3D scan as a 2D range image, and the search for features and correspondence analysis is restricted to the 2D space. This significantly reduces the computational time, however the transformation to the curvelet domain to find suitable features, computation of feature descriptors and the rejection of false correspondences add to the run-time.

\begin{center}
	\begin{table}[htb!]
	\caption{Average run-times (s) for pair-wise scan registration for each data-set used for comparisons.}
	\newcolumntype{L}{>{\centering\arraybackslash}m{2.80cm}}
	\newcolumntype{B}{>{\centering\arraybackslash}m{3.75cm}}
	\newcolumntype{C}{>{\centering\arraybackslash}m{6.25cm}}
	\renewcommand{\arraystretch}{1.2}% Wider
    \begin{tabular}{ |C|L|L|L| }
  	\hline
  	 & \textbf{FPFH + SAC-IA} & \textbf{Harris-3D + FPFH} & \textbf{Proposed Method}   \\
  	 \hline
  	 \textbf{Indoor Mars-dome data-set (s)} & 2745.59 & 4307.76 & 91.15  \\
  	\hline
  	\textbf{Outdoor simulated CSA Mars-Yard (s)} & 1049.09 & 1076.62 & 101.1 \\
  	\hline
  	\textbf{Outdoor real-world CSA Mars-Yard (s)} & 138.45  & 170.54 & 107.04\\
  	\hline
  	\end{tabular}
  	\label{table:run_times}
    \end{table}  
\end{center}

%%%%%%%%%%%%%%%%%%%%%%%%%%%%%%%%%%%%%%%%%%%%%%%%%%%%%%%%%%%%
\section{Conclusions and Future Work}
In this work, a curvelet transform based method for improving the alignment accuracy of standard registration algorithms is presented. Suitable features in the curvelet domain are found via difference of curvelets operator at multiple scales. Curvelet features help in obtaining a higher level understanding of the environment by extracting regions of interest, thereby reducing the amount of data to be processed. The neighborhood around the feature is captured by a shape descriptor computed from spatial histograms of image gradients, which is further used in establishing correspondences using nearest neighbor matching between scan points from different viewpoints. RANSAC based filtering of feature correspondences is followed by SVD based alignment of the laser scans. The proposed method is verified experimentally by first evaluating the scan registration accuracy on a publicly available indoor UTIAS Mars-dome data-set, as well as simulated and real-world CSA Mars-yard data-sets. The sparse-featured, unstructured Mars-like terrain poses a significant challenge for global registration methods relying on features such as FPFH and Harris-3D. Curvelet feature based scan registration is shown to produce lower translation and rotation errors as compared with other methods. Second, the resulting rover trajectory is generated by integrating scans after incremental pair-wise scan registration for outdoor data-sets and the proposed method is shown to produce lower drift. Third the quality of the scan registration is verified by integrating scan points after registration into a final map. Visually, it is shown that the map generated from the proposed method is sharper as compared with the maps generated from FPFH and Harris-3D based methods.  

Although curvelet features efficiently localize regions with sharp discontinuities along the curve, due to the compact representation of the underlying surface structure as point features, a lot of geometric information, particularly in the smoothly varying regions is lost during the extraction process, the inclusion of which could potentially result in higher registration accuracy. In addition, a better understanding of the underlying surface structure, as well as the overall information content within the scan is needed. In this paper, we restrict our analysis to the effects of the selected global methods on the scan registration accuracy in challenging outdoor data-sets. However, to the best knowledge of the authors, a detailed analysis of the repeatability of various 3D features under view-point variations and robustness with respect to noise, scale changes, rotations, and translations in planetary environments is lacking in the literature. It is definitely a sizable undertaking, and we have therefore left this as an area for future work.

The range-imaging step of the curvelet algorithm is susceptible to the quantization errors during the projection of laser points onto the range image plane, caused by non-uniform sampling by various laser sensors along the azimuth and elevation direction. Pixels in the range image could correspond to more than one range measurement or none at all, resulting in information loss which may negatively affect the scan registration accuracy in cluttered indoor environments. Gaussian filtering with a small $3 \times 3$ kernel has been employed to reduce the noise in the range-intensities during the range imaging step. The effects of surface de-noising on the scan registration accuracy has been left as an area for future work.

Identification of dynamic objects in the scene would further aid the scan registration accuracy in real-world settings. Inclusion of IMU data in the registration pipeline could provide an initial estimate for the optimizers in the scan registration methods, which could result in faster convergence and more accurate registration. Global optimization as a post-processing step will further increase the scan registration accuracy, and is proposed as a future extension of the work.

\subsection*{\textbf{Funding}}
This work was supported in part by the Natural Sciences and Engineering Research Council (NSERC) of Canada through the NSERC Canadian Field Robotics Network (SPG-2012-NCRFN).

\bibliographystyle{apalike}
\bibliography{jfrExampleRefs}

\end{document}